%% file: main.tex
\documentclass[10pt,journal,compsoc]{IEEEtran}

\usepackage{caption}
\usepackage{multirow}
\usepackage{booktabs}
\usepackage{graphicx}
\usepackage{pdflscape}
\usepackage{rotating}
\usepackage{color}
\usepackage{microtype}
\usepackage{amssymb}
\usepackage{amsfonts}
\usepackage{bbding}
\usepackage{pifont}
\usepackage[ruled,linesnumbered]{algorithm2e}
\usepackage[pagebackref=false,breaklinks=true,letterpaper=true,colorlinks,bookmarks=false]{hyperref}
\newcommand{\revision}[1]{#1}

\newlength\savewidth

\begin{document}
\title{3D Object Detection from Images for Autonomous Driving: A Survey}

\author{Xinzhu Ma,
        Wanli Ouyang,
        Andrea Simonelli,
        Elisa Ricci
\IEEEcompsocitemizethanks{\IEEEcompsocthanksitem Xinzhu Ma and Wanli Ouyang are with Shanghai AI Laboratory and with the Chinese University of Hong Kong.
Email: maxinzhu@pjlab.org.cn, ouyangwanli@pjlab.org.cn. This work is partially supported by the National Key R\&D Program of China (NO.2022ZD0160101).
\IEEEcompsocthanksitem Andrea Simonelli and Elisa Ricci are with the Department of Information Engineering and Computer Science (DISI) at the University of Trento and with Fondazione Bruno Kessler (FBK).
Email: andrea.simonelli@unitn.it, 
e.ricci@unitn.it
}}

\markboth{Journal of \LaTeX\ Class Files,~Vol.~14, No.~8, August~2015}%
{Shell \MakeLowercase{\textit{et al.}}: Bare Advanced Demo of IEEEtran.cls for IEEE Computer Society Journals}

\input{components/abstract}
\input{components/introduction}

\input{components/task}
\input{components/dataset}
\input{components/frameworks}
\input{components/components}
\input{components/aux_data}
\input{components/future_direction}
\input{components/conclusion}

\ifCLASSOPTIONcaptionsoff
  \newpage
\fi

\bibliographystyle{IEEEtran}
\bibliography{references}

\begin{IEEEbiography}[{\includegraphics[width=1in,height=1.25in,clip,keepaspectratio]{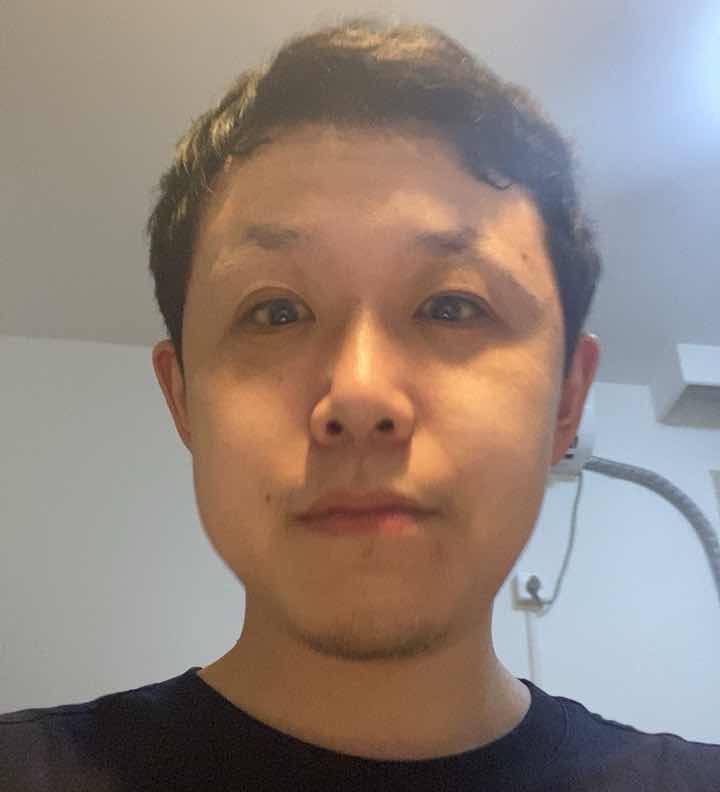}}]
{Xinzhu Ma} received his B.Eng and M.P's degree from Dalian University of Technology in 2017 and 2019, respectively. After that, He got the Ph.D degree from the University of Sydney in 2023. He 
is currently a postdoctoral researcher at the Chinese University of Hong Kong. His research interests include deep learning and computer vision.
\end{IEEEbiography}

\begin{IEEEbiography}[{\includegraphics[width=1in,height=1.25in,clip,keepaspectratio]{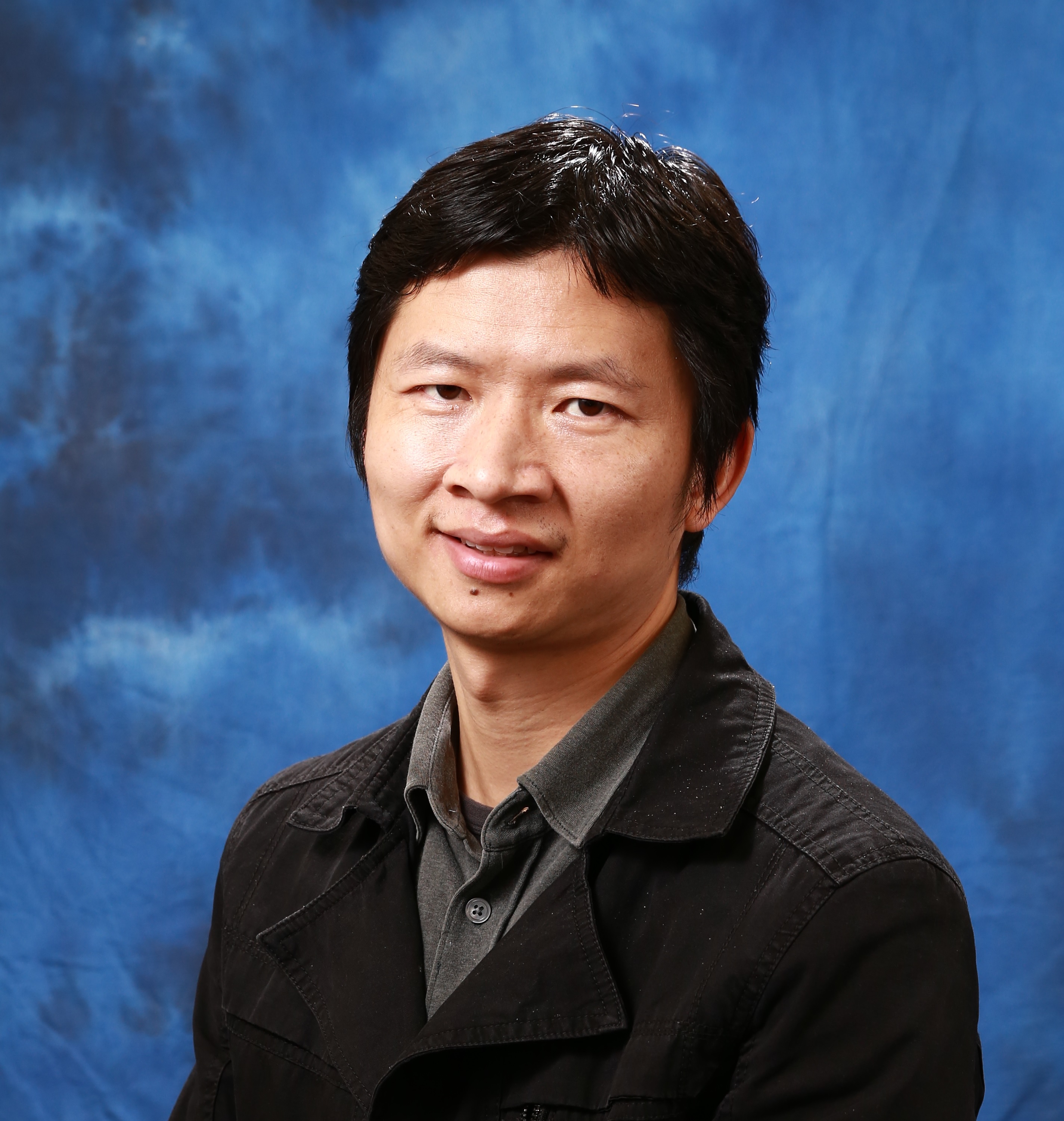}}]{Wanli Ouyang} received the Ph.D degree in the Department of Electronic Engineering, The Chinese University of Hong Kong. After that, He joined the University of Sydney as a senior lecturer. He is now a professor at the Chinese University of Hong Kong. His research interests include image processing, computer vision and pattern recognition. He is a senior member of IEEE.
\end{IEEEbiography}

\begin{IEEEbiography}[{\includegraphics[width=1in,height=1.25in,clip,keepaspectratio]{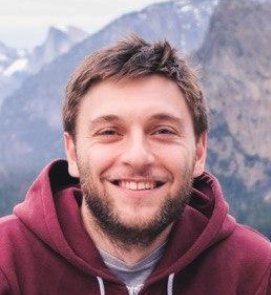}}]{Andrea Simonelli} received his master’s degree in 2018 from the University of Trento, Italy, where he is now a PhD student in Information and Communication Technologies. His research grant is co-funded by Fondazione Bruno Kessler and Mapillary Research. His main research interests are deep learning and computer vision.
\end{IEEEbiography}

\begin{IEEEbiography}[{\includegraphics[width=1in,height=1.25in,clip,keepaspectratio]{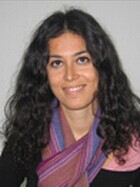}}]{Elisa Ricci} received the Ph.D degree from the University of Perugia in 2008. She is an associate professor at the University of Trento and a researcher at Fondazione Bruno Kessler. She has since been a post-doctoral researcher at Idiap, Martigny, and Fondazione Bruno Kessler, Trento. She was also a visiting researcher at the University of Bristol. Her research interests are mainly in the areas of computer vision and machine learning. She is a member of the IEEE.
\end{IEEEbiography}

\newpage
\input{components/appendix}

\end{document}

%% file: components/abstract.tex
\IEEEtitleabstractindextext{%
\begin{abstract}
3D object detection from images, one of the fundamental and challenging problems in autonomous driving, has received increasing attention from both industry and academia in recent years. Benefiting from the rapid development of deep learning technologies, image-based 3D detection has achieved remarkable progress. Particularly, more than 200 works have studied this problem from 2015 to 2021, encompassing a broad spectrum of theories, algorithms, and applications. However, to date no recent survey exists to collect and organize this knowledge. In this paper, we fill this gap in the literature and provide the first comprehensive survey of this novel and continuously growing research field, summarizing the most commonly used pipelines for image-based 3D detection and deeply analyzing each of their components. Additionally, we also propose two new taxonomies to organize the state-of-the-art methods into different categories, with the intent of providing a more systematic review of existing methods and facilitating fair comparisons with future works. In retrospect of what has been achieved so far, we also analyze the current challenges in the field and discuss future directions for image-based 3D detection research.
\end{abstract}

}

\maketitle

%% file: components/introduction.tex
\IEEEdisplaynontitleabstractindextext
\IEEEpeerreviewmaketitle

\ifCLASSOPTIONcompsoc
\IEEEraisesectionheading{\section{Introduction}\label{sec:introduction}}
\else
\section{Introduction}
\label{sec:introduction}
\fi

\IEEEPARstart{A}UTONOMOUS driving has the potential to radically change people's lives, improving mobility and reducing travel time, energy consumption, and emissions. Therefore, unsurprisingly, in the last decade both research and industry have put significant efforts to develop self-driving vehicles. As one of the key enabling technologies for autonomous driving, 3D object detection has received a lot of attention, and deep learning-based 3D object detection approaches have gained popularity in recent years.

Existing 3D object detection approaches can be roughly categorized into two groups according to whether the input data are images or LiDAR signals (generally represented as point clouds). Although the LiDAR-based methods show promising performances, the expensive and cumbersome sensors restrict the application of these algorithms. Besides, the intrinsic properties of LiDAR sensors also determine that they are difficult to cover some corner cases, {\it e.g.} long-range objects.
Meanwhile, the methods based on the cheaper and easy-to-deploy cameras show great potential in many scenarios. Consequently, although estimating 3D bounding boxes from images only faces a much greater challenge, this task still draws lots of attention and gradually becomes a hot topic in the computer vision (CV) community. In the past six years, more than 80 papers have been published on top-tier conferences and journals in this area, achieving several breakthroughs both in terms of detection accuracy and inference speed. 

\revision{However, previous surveys of 3D object detection, such as \cite{survey_ad,survey3d}, generally focus more on the LiDAR-based models, instead of the image-based ones, thus can not provide a clear and detailed presentation of the image-based models for the public.}
In this paper, we provide the first comprehensive and structured review of the recent advances in image-based 3D object detection based on deep learning techniques. In particular, this survey summarizes the previous research works in this area, ranging from the pioneering methods \cite{3dop,mono3d} to the most recent approaches~\cite{monodistill,weakm3d} published in ICLR'2022. The survey reviews and analyses both the high-level frameworks and the specific design choices of each required component for an image-based 3D detection model ({\it e.g.} feature extraction, loss formulation, etc.). Furthermore, we propose two novel taxonomies to categorize existing methods, {\it i.e.} in terms of their adopted frameworks and of the used input data. This is intended to facilitate both the systematic analysis of current approaches and a fair comparison in performance for future works.

The main contributions of this work can be summarized as follows:
\begin{enumerate}
\item To the best of our knowledge, this is the ﬁrst work that surveys image-based 3D detection methods for autonomous driving. 80+ image-based 3D detectors and 200+ related research works are reviewed.
\item We provide a comprehensive review and an insightful analysis on the key aspects of the problem, including datasets, detection pipelines, technical details, etc.
\item We propose two novel taxonomies of the state-of-the-art methods, with the purpose of helping the readers to easily acquire knowledge on this new and growing research field.
\item We summarize the main issues and future challenges in image-based 3D detection, outlining some potential research directions for future work.
\end{enumerate}

\subsection{Scope}
Image-based 3D object detection is closely related to many other tasks, such as 2D object detection, depth estimation, stereo matching, LiDAR-based 3D object detection, etc. It is impractical to review these related technologies in detail in a single manuscript. In this work, we mainly focus on image-based 3D detection, and only some representative methods of the related fields are introduced.

This paper mainly focuses on the major progress of the last six years, especially the works published in top-tier conferences and journals. In addition to the technical details of these image-based 3D detectors,  taxonomies, datasets, evaluation metrics, and potential challenges/research directions are also presented in this survey. Additionally, we also provide a continuously maintained project page on: \url{https://github.com/xinzhuma/3dodi-survey}.

\subsection{Organization of the Manuscript}
The manuscript is organized as follows. A brief introduction to the image-based 3D detection task is given in Section~\ref{sec:task}. The commonly used datasets and metrics are summarized in Section~\ref{sec:dataset}. Section~\ref{sec:frameworks} describes the common frameworks. These three sections are intended for beginners in order for them to quickly acquire a good understanding of the problem of image-based 3D detection. In Section~\ref{sec:components} we compare the details of each component typically required in a 3D detector, while Section~\ref{sec:aux_data} discusses commonly used auxiliary input data. These two sections can help the researchers in this field to build a clear, in-depth and more structured knowledge of the topic. In Section~\ref{sec:future}, we point out some possible future research directions, which can provide insights for future works. Finally, conclusions are drawn in Section ~\ref{sec:conclusion}.

%% file: components/task.tex
\section{Task}
\label{sec:task}

\begin{figure}[t]
\centering
\includegraphics[width=\columnwidth]{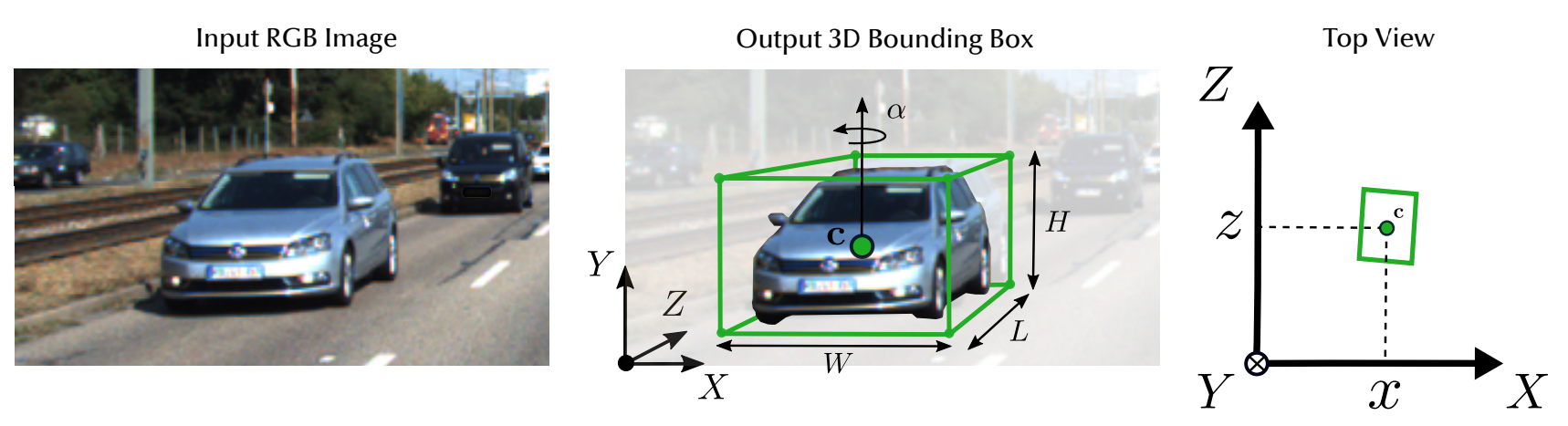}
\vspace{-3pt}
\caption{Illustration of the monocular 3D object detection task. Given an input image ({\it left}), it aims to predict a 3D bounding box (represented by its location $(x,y,z)$, dimension $(h,w,l)$, and orientation $\theta$ for each object ({\it middle}). We also show the bird's eye view for better visualization ({\it right}).}
\label{fig:task}
\vspace{-3pt}
\end{figure}

Given the RGB images and corresponding camera parameters, the goal of image-based 3D object detection is to classify and localize the objects of interest. Each object is represented by its category and bounding box (denoted as B for short) in the 3D world space.  Generally, the 3D bounding box is parameterized by its location $[x, y, z]$, dimension $[h, w, l]$, and orientation $[\theta, \phi, \psi]$\footnote[1]{Location and orientation are also called translation and rotation in some works.} relative to a predefined reference coordinate system ({\it e.g.} the one of the ego-vehicle which recorded the data). In most autonomous driving scenarios, only the heading angle $\theta$ around the up-axis (yaw angle) is considered.
Figure~\ref{fig:task} visualizes an  example result on both the 2D image plane and the bird's eye view.

While the general problem of the image-based 3D object detection can be stated as described above, it is worth mentioning that: i): besides the category and 3D bounding boxes, some benchmarks require additional predictions, {\it e.g.} 2D bounding box for the KITTI dataset \cite{kitti} and the velocity/attribute for the nuScenes dataset \cite{nuscenes}. ii): although only images and camera parameters are initially provided for this task, the adoption of auxiliary data ({\it e.g.} stereo pairs, LiDAR signals, etc.) is common in this field.

%% file: components/dataset.tex
\begin{table*}[!t]
\centering
\caption{The summary of datasets that can be used for image-based 3D object detection in autonomous driving scenarios. Some datasets are proposed for multiple tasks, and here we report the numbers for the 3D detection benchmark ({\it e.g.} KITTI 3D released more than 40K images, and about 15K of them are used for 3D detection). $^{\dagger}$: not released.}
\resizebox{\linewidth}{!}{
\begin{tabular}{rcrrrrcccccccc}
\toprule
\multirow{2}{*}{Dataset} & \multirow{2}{*}{Year}  & \multicolumn{4}{c}{Size} & \multicolumn{3}{c}{Diversity} & \multicolumn{3}{c}{Additional data}&\multirow{2}{*}{Benchmark}\\
\cmidrule(lr){3-6}  \cmidrule(lr){7-9} \cmidrule(lr){10-12} 
 ~ & ~ & \# Train & \# Val & \# Test & \# Boxes & \# Scenes & \# Classes$^*$ &  Night/Rain & Stereo & Temporal & LiDAR & ~\\
\midrule
KITTI 3D~\cite{kitti} & 
2017 & 7,418$\times$1 & - & 7,518$\times$1 & 200K & - & 3 & No/No & Yes & Yes & Yes & Yes \\
ApolloCar3D \cite{apollocar3d} & 2019 & 4,036$\times$1 & 200$\times$1 & 1,041$\times$1 & 60K & - & 1 & Yes/No & Yes & Yes & Yes & Yes \\
Argoverse~\cite{argoverse} & 
2019 & 39,384$\times$7 & 15,062$\times$7 & 12,507$\times$7 & 993K & 113 & 15 & Yes/Yes  & Yes & Yes & Yes & Yes\\
Lyft L5~\cite{lyft}& 
2019 & 22,690$\times$6 & - & 27,468$\times$6 & 1.3M & 366 & 9 & No/No  & No & Yes & Yes & No\\
H3D~\cite{h3d}& 
2019 &  8,873$\times$3 & 5,170$\times$3 & 13,678$\times$3 & 1.1M & 160 & 8 &  No/No  & No & Yes & Yes & No\\
A*3D$^{\dagger}$~\cite{a*3d}& 2019 & 39,179$\times$1 & - & - & 230K & - & 7 & Yes/Yes  & Yes & - & Yes & No\\
nuScenes~\cite{nuscenes} & 
2019 &  28,130$\times$6 & 6,019$\times$6 & 6,008$\times$6 & 1.4M & 1,000 & 10 & Yes/Yes  & No & Yes & Yes & Yes\\
Waymo Open~\cite{waymo} & 
2019 & 122,200$\times$5 & 30,407$\times$5 & 40,077$\times$5 & 12M & 1,150 & 3 & Yes/Yes  & No & Yes & Yes & Yes\\
CityScapes 3D~\cite{cityscapes3d}& 
2020 & 2,975$\times$1 & 500$\times$1 & 1,525$\times$1 & 40K &  - & 6 & Yes/Yes & Yes & No & No & Yes\\
A2D2 \cite{a2d2} &
2020 & 12,497$\times$1 & - & - &  - & - & 14 & No/Yes & No & Yes & Yes & No \\
KITTI-360 ~\cite{kitti360}& 2021 & - & - & - & 68K & - & 2 & No/No & Yes & Yes & Yes & Yes\\
\bottomrule
\end{tabular}}
\label{table:summary_of_dataset}
\vspace{-3pt}
\end{table*}

\section{Datasets and Evaluation}
\label{sec:dataset}

\subsection{Datasets}
It is a well known fact that the availability of large-scale datasets is essential for the success of the data-driven deep learning techniques. For image-based 3D object detection in autonomous driving scenario, the main characteristics of the publicly available datasets \cite{kitti,nuscenes,apollocar3d,argoverse,lyft,h3d,a*3d,waymo,cityscapes3d,a2d2,kitti360} are summarized in Table~\ref{table:summary_of_dataset}. Among these datasets, the KITTI 3D \cite{kitti}, nuScenes \cite{nuscenes}, and Waymo Open \cite{waymo} are the most commonly used and greatly promote the development of 3D detection. In the following, we provide the main information about these benchmarks.

\noindent
{\bf Basic information.}
For most of the past decade, KITTI 3D was the only dataset to support the development of image-based 3D detectors.  KITTI 3D provides front-view images with a resolution of $1280\times384$ pixels. In 2019 the nuScenes and Waymo Open datasets were introduced. In the nuScenes dataset six cameras are used to generate 360$^{\circ}$ view with a resolution of $1600\times 900$ pixels. Similarly, Waymo Open also captures 360$^{\circ}$ view using five synchronized cameras, and the resolution of image is $1920\times1280$ pixels.

\noindent
{\bf Dataset size.}
The KITTI 3D dataset provides 7,481 images for training and 7,518 images for testing, and it is common practice to split the training data into a training set and a validation set \cite{3dvp,3dop,demystifying}. As the most commonly used one, 3DOP's split \cite{3dop} includes 3,712 and 3,769 images for training and validation, respectively. The large-scale nuScenes and Waymo Open provide about 40K and 200K annotated frames, and use multiple cameras to capture the panoramic view of each frame. In particular, nuScenes provides 28,130 frames, 6,019 frames, and 6,008 frames for training, validation, and testing (six images per frame). Waymo Open, the largest one, gives 122,200 frames for training, 30,407 frames for validation, and 40,077 frames for testing (five images per frame). It is worth mentioning that both these two datasets collect about 1.4M frames raw data, while Waymo Open annotates them at a 5$\times$ higher frequency than nuScenes.  Besides, all the three datasets only release the annotations for training/validation set, and the evaluation on test set can only be conducted on their official testing servers.

Note that most papers only use the KITTI 3D dataset (with a focus on the {\it Car} category) for evaluation, except for the works in \cite{monodis_journal,centernet, 3dgck, caddn, fcos3d, monoef,dd3d,geo_augmentation,pgd,detr3d} reporting performances on nuScenes or Waymo Open.
Nevertheless, in future works, evaluating on these large-scale datasets is essential for assessing the effectiveness of the algorithms.


\noindent
{\bf Diversity.}
The KITTI 3D dataset is captured in Karlsruhe, Germany in the daylight and good weather conditions. It mainly evaluates objects from three categories (Car, Pedestrian, and Cyclist), and divides them into three difficulty levels according to the height of 2D bounding boxes, occlusion, and truncation. The nuScenes dataset consists of 1000 scenes of 20s captured in Boston and Singapore. Differently from the KITTI 3D benchmark, these scenes have been captured at different times of the day (including night) and in different weather conditions ({\it e.g.} rainy day). There are ten categories of objects for the 3D detection, and nuScenes also annotates the attribute labels for each category, {\it e.g.} moving or parked for a car, with or without a rider for a bicycle. These attributes can be regarded as fine-grained class labels, and the accuracy of attribute recognition is also considered in the nuScenes benchmark. 
The Waymo Open dataset covers 1,150 scenes, shot in Phoenix, Mountain View, and San Francisco under multiple weather conditions, including night and rainy days. Similar to KITTI 3D, Waymo Open also deﬁnes two difﬁculty levels for the 3D detection task according to the number of LiDAR points contained in 3D bounding boxes. The objects of interest in its benchmark include vehicles, pedestrians, and cyclists.

\noindent
{\bf Additional data.}
In addition to the RGB images and the corresponding camera parameters, these datasets also provide additional data that can be optionally used in the image-based 3D detection task. Specifically, all the three datasets provide the LiDAR signals and temporally preceding frames (note that these preceding images may be \emph{unlabelled} because these datasets only annotate the key-frames from the collected videos), and the KITTI 3D dataset also provides the stereo pairs to support 3D object detection from stereo images. Section~\ref{sec:aux_data} discusses the use of these auxiliary data in existing methods in detail.

\noindent
{\bf Evaluation metric.}
The evaluation metric for the 3D object detection is built on the Average Precision (AP) ~\cite{ap,voc}, and we recommend readers refer to Appendix \ref{supp:metric} for the review of original AP metric and the introduction of its variants adopted in the 3D detection task.

\begin{figure}
\centering 
\resizebox{\linewidth}{!}{
\includegraphics{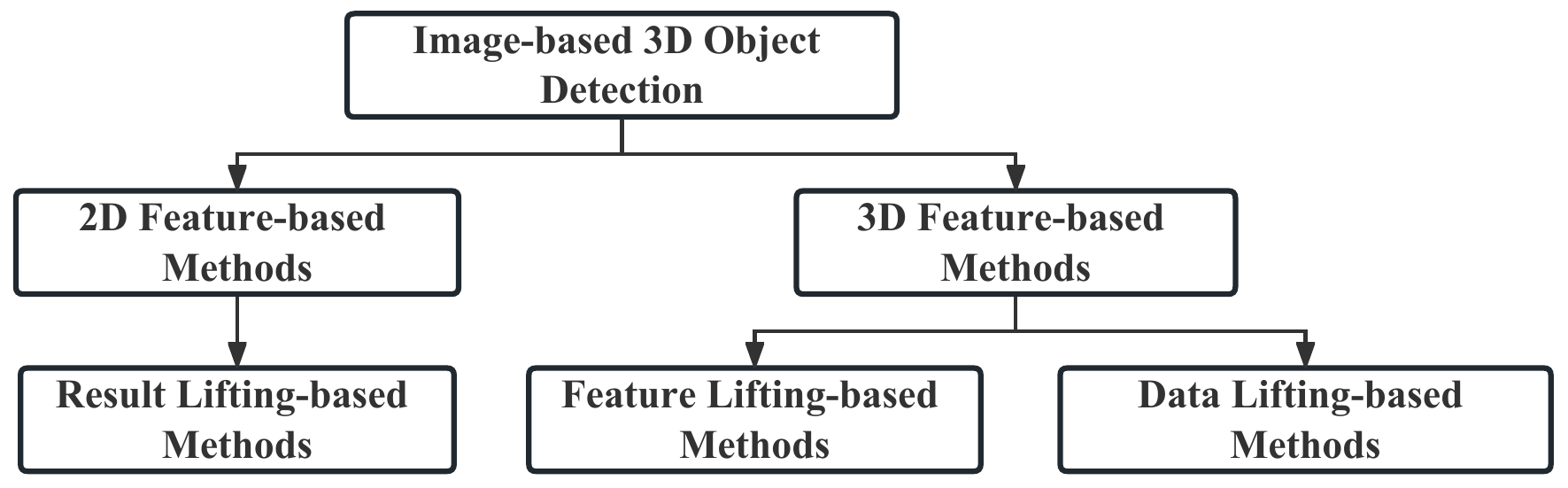}}
\caption{The proposed taxonomy for image-based 3D detection. The methods are first divided into `2D feature-based methods' and `3D feature-based methods'. Then we further group them into `result-lifting-based methods', `feature-lifting-based methods', and `data-lifting-based methods'.}
\label{fig:taxonomy}
\vspace{-8pt}
\end{figure}

%% file: components/frameworks.tex
\begin{figure*}
\begin{center}
\resizebox{\linewidth}{!}{
\includegraphics{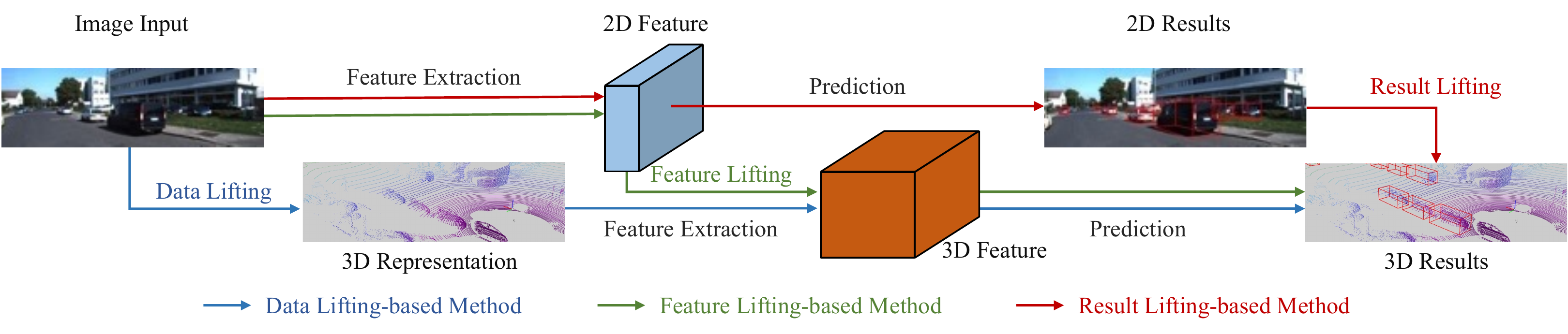}}
\end{center}
\vspace{-10pt}
\caption{Illustration of the image-based 3D object detection pipelines. We show the data flows of data-lifting, feature-lifting, and result-lifting methods with \textcolor[RGB]{51,112,204}{blue}, \textcolor[RGB]{77,179,77}{green}, and \textcolor[RGB]{213,43,43}{red} arrows respectively.}
\label{fig:pipelines}
\vspace{-5pt}
\end{figure*}

\section{Frameworks}
\label{sec:frameworks}

In this section, we summarize the image-based 3D detection methods in terms of the high-level paradigm. Specifically, we first introduce a new taxonomy for this task, and then discuss the existing methods accordingly.

\subsection{Taxonomy}
\begin{figure*}
\begin{center}
\resizebox{\linewidth}{!}{
\includegraphics{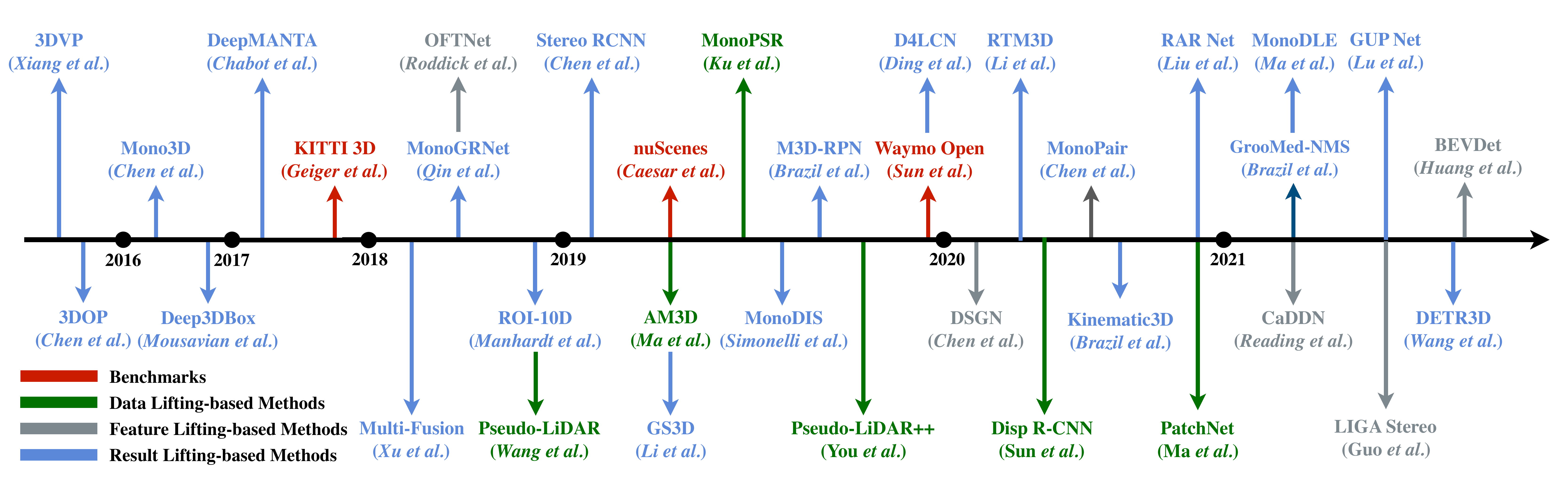}}
\end{center}
\vspace{-10pt}
\caption{Chronological overview of the most relevant image-based 3D detection methods and the profound benchmarks.}
\label{fig:milestones}
\vspace{-8pt}
\end{figure*}

As shown in Figure \ref{fig:taxonomy}, we propose to group the image-based 3D detectors into two branches: (i) \emph{the methods based on 2D features}, and (ii) \emph{the methods based on 3D features}. We believe this taxonomy can help beginners quickly establish a preliminary understanding of the existing methods. Furthermore, our taxonomy further divides these methods into (i) \emph{the methods based on result lifting}, (ii) \emph{the methods based on feature lifting}, and (iii) \emph{the methods based on data lifting}, which indicates the core problem of image-based 3D object detection: how to generate 3D results from 2D data. Particularly, the result lifting-based methods first estimate the 2D locations (and other items such as orientation, depth, etc.) of the objects in the image plane from the 2D features, and then lift the 2D detections into the 3D space. The feature lifting-based methods generate the 3D features by lifting the 2D features and then predict the final results in the 3D space. 
Similarly, the data lifting-based methods can also generate the 3D results directly, but they lift the input data from 2D to 3D, instead of the features. Figure \ref{fig:pipelines} compares the data flows of these detection pipelines.
According to the aforementioned taxonomy, we highlight the milestone methods (with the key benchmarks) in Figure \ref{fig:milestones}.

Because there is no specific taxonomy for image-based 3D detection, previous works generally adopt the classic 2D detection taxonomy to divide the object 3D detectors into single-shot (one-stage) methods and region-based (two-stage) methods. We argue our taxonomy is more suitable for image-based 3D detection because: (i) Our taxonomy groups the methods based on the feature representations, the foundation of the deep learning-based methods, thus it can help the readers build a structured knowledge quickly. (ii) Our taxonomy indicates how a detector aligns the dimension mismatch between the 2D input data and the 3D results, which is the core problem of this task. (iii) Our taxonomy can clearly define the existing methods, while the previous ones can not. {\it E.g.}, the pseudo-LiDAR-based methods (will be introduced in Section \ref{sec:framework_pseudolidar}) can adopt any LiDAR-based detectors, including region-based methods and single-shot methods.  Therefore, it is hard to assign these methods to either side.

\subsection{Methods Based on 2D Features}
The first group is the `methods based on 2D features'. Given an input image, they first estimate the 2D locations, orientations, and dimensions (see Figure \ref{fig:task} for the visualization of these items) from the 2D features, and then recover the 3D locations from these intermediate results. Therefore, these methods can also be called 'result lifting-based methods'.  In particular, to get an object's 3D location $[x, y, z]$, an intuitive and commonly used solution is to estimate the depth value $d$ using CNNs, and then lift the 2D projection into the 3D space using:
\begin{equation}
\label{equ:2dto3d}
\left\{
      \begin{array}{lr}
      \vspace{4pt}
      z = d ,\\
      \vspace{4pt}
      x = (u-C_{x}) \times z / f  ,\\
      y = (v-C_{y}) \times z / f  ,\\
      \end{array}
\right.
\end{equation}
where $(C_{x}, C_{y})$ is the principal point, $f$ is the focal length, and  $(u, v)$ is object's 2D location. Also note, these methods only need the depths of the center of the objects, which are different from the methods which require dense depth maps, {\it e.g.} Pseudo-LiDAR \cite{pseudolidar}. Furthermore, because the 2D feature based methods are similar to 2D detectors in the overall framework, we introduce these works with classic taxonomy used in 2D detection field, {\it i.e.} region-based methods and single-shot methods, for better presentation.

\subsubsection{Region-based Methods}
The region-based methods follow the high-level idea of the R-CNN series~\cite{rcnn,fastrcnn,fasterrcnn} in 2D object detection. In this framework, after generating category-independent region proposals from input images, features are extracted from these regions by CNNs \cite{fasterrcnn,maskrcnn}. Finally, R-CNN uses these features to further refine the proposals and determine their category labels. Here we summarize the novel designs of the region-based framework for the image-based 3D detection.

\noindent
{\bf Proposal generation.}
Different from the commonly used proposal generation methods \cite{ss_journal,edgeboxes} in the 2D detection field, a simple method to generate proposals for 3D detection is to tile the 3D anchors ({\it shape templates of the proposals}) in the ground plane and then project them to the image plane as proposals. However, this design generally leads to a huge computational overhead. To reduce the searching space, Chen {\it et al.} \cite{mono3d,3dop,3dop_journal} proposed the pioneering Mono3D and 3DOP by removing the proposals with low confidence using domain-specific priors ({\it e.g. shape, height, location distribution, etc}) for monocular and stereo based methods respectively. Besides, Qin {\it et al.} \cite{tlnet} proposed another scheme which estimates an objectness confidence map in the 2D front-view, and only the potential anchors with high objectness confidence are considered in the subsequent steps. In summary, 3DOP~\cite{3dop} and Mono3D~\cite{mono3d} compute confidence of proposals using geometric priors, while Qin \emph{et al.} \cite{tlnet} uses a network to predict the confidence map.

With the Region Proposal Network (RPN) \cite{fasterrcnn}, the detectors can generate 2D proposals using features from the last shared convolutional layer instead of external algorithms, which saves most of the computational cost, and lots of image-based 3D detectors \cite{shiftrcnn,keypointsgeo,multifusion,monofenet,monodis,monodis_journal,stereorcnn,centervoting,monorcnn,ida3d,tlnet,disprcnn,zoomnet,oc_stereo,car_parsing} adopted this design.

\noindent
{\bf Introducing spatial information. }
Chen {\it et al.} \cite{stereorcnn} extended the design of RPN and R-CNN combination to the stereo based 3D detection. They proposed to extract the features from left image and right image separately and used the fused feature to generate proposals and predict the final results. This design allows the CNN to implicitly learn disparity/depth cues from stereo pairs, and is adopted by the following stereo based 3D detectors \cite{ida3d,dispnet,zoomnet}. Also for the same purpose of providing depth information, Xu and Chen \cite{multifusion} proposed another scheme, Multi-Fusion, for monocular 3D detection. In particular, they first generate depth maps for input images using an off-the-shelf depth estimator~\cite{monodepth,dorn}, and then design a region-based detector with multiple information fusion strategies for the RGB images and depth maps. It is worth noting that the strategy of providing depth cues with extra depth estimator for monocular images is embraced by several works \cite{roi10d,pseudolidar,am3d,monowithpl,d4lcn,ddmp3d,decoupled3d,pseudolidar,patchnet,neighborvote,da3ddet}. Nevertheless, Stereo R-CNN \cite{stereorcnn} and Multi-Fusion \cite{multifusion} are similar in the high-level paradigm based on the fact that they both adopt the region-based framework and introduce another image (or map) to provide the spatial cues.

\subsubsection{Single-Shot Methods}
The single-shot object detectors directly predict class probabilities and regress other items of the 3D boxes from each feature position. As a consequence, these methods generally have faster inference speed than the region-based methods, which is vital in the context of autonomous driving. The use of only CNN layers in single-shot methods also facilitate their deployment on different hardware architectures. Besides, some relevant works \cite{centernet,fcos3d,focalloss,fcos} showed that the single-shot detectors can also achieve promising performance. Based on the above reasons, lots of recent methods adopted this framework.

\noindent
{\bf Basic single-shot models.}
Currently, there are two single-shot prototypes used in image-based 3D detection. The first one is anchor-based, proposed by Brazil and Liu~\cite{m3drpn}. In particular, this detector is essentially a tailored RPN for monocular 3D detection, and it generates both 2D anchors and 3D anchors for the given images. Different from the category-independent 2D anchors, the shape of 3D anchors generally have a strong correlation to their semantic label, {\it e.g.} an anchor with a shape of `$1.5m\times 1.6m \times 3.5m$' is usually a car rather than a pedestrian. Therefore, this 3D RPN can be used as the single-shot 3D detector and has been adopted by several methods \cite{d4lcn,kinematic3d,groomednms,ddmp3d}.

Besides, in 2019, Zhou {\it et al.} ~\cite{centernet} proposed an anchor-free single-shot detector named CenterNet, and extended it to image-based 3D detection. In particular, this framework encodes the object as a single point (the center point of the object) and uses key-point estimation to ﬁnd it. Besides, several parallel heads are used to estimate the other properties of the object, including depth, dimension, location, and orientation. Although this detector seems very simple in architecture, it achieves promising performance across several tasks and datasets. Later on, many following works \cite{fcos3d,monoef,monopair,monodle,smoke,rtm3d,gupnet,m3dssd,monoflex,autoshape, movi3d} adopted this design.

\noindent
{\bf Extensions.} There are lots of improvements based on these two prototypes, and we summarize them in Section \ref{sec:components} due to their modular design.

\subsection{Methods Based on 3D Features}
Another branch of the proposed taxonomy is the `methods based on 3D features'. The main feature of these methods is they first generate the 3D features from the images, and then directly estimate all items of the 3D bounding boxes, including the 3D locations, in the 3D space. According to how to get the 3D features, we further group these methods into 'feature lifting-based methods' and 'data lifting-based methods'.

\subsubsection{Feature Lifting-based Methods}
\label{sec:framework_feature_trans}
The general idea of the feature lifting-based methods is to transform the 2D image features in the image coordinate system into the 3D voxel features in the world coordinate system. Moreover, existing feature lifting-based methods \cite{oftnet,dsgn,caddn,ligastereo,bevdet} further collapse the 3D voxel features along the vertical dimension, corresponding to the height of objects, to generate the Bird's Eye View (BEV) features before estimating final results. For this kind of methods, the key problem is how to transform the 2D image features into the 3D voxel features.  We discuss this problem in the following.

\begin{figure}
\begin{center}
\includegraphics[width=0.48\linewidth]{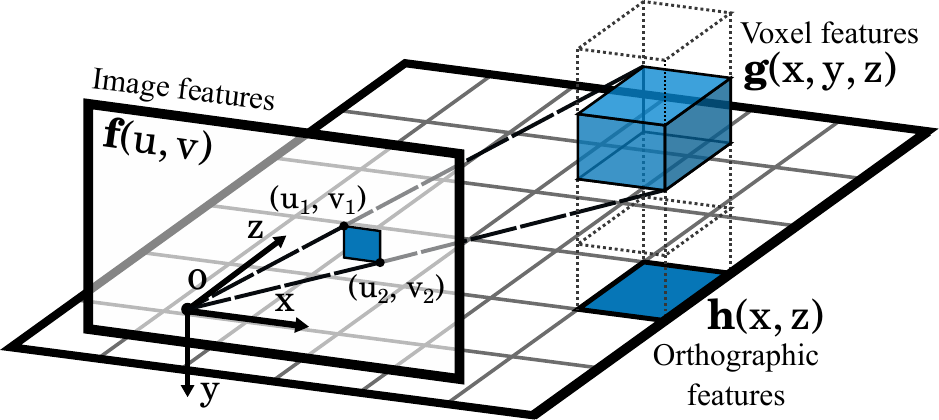}  
\includegraphics[width=0.48\linewidth]{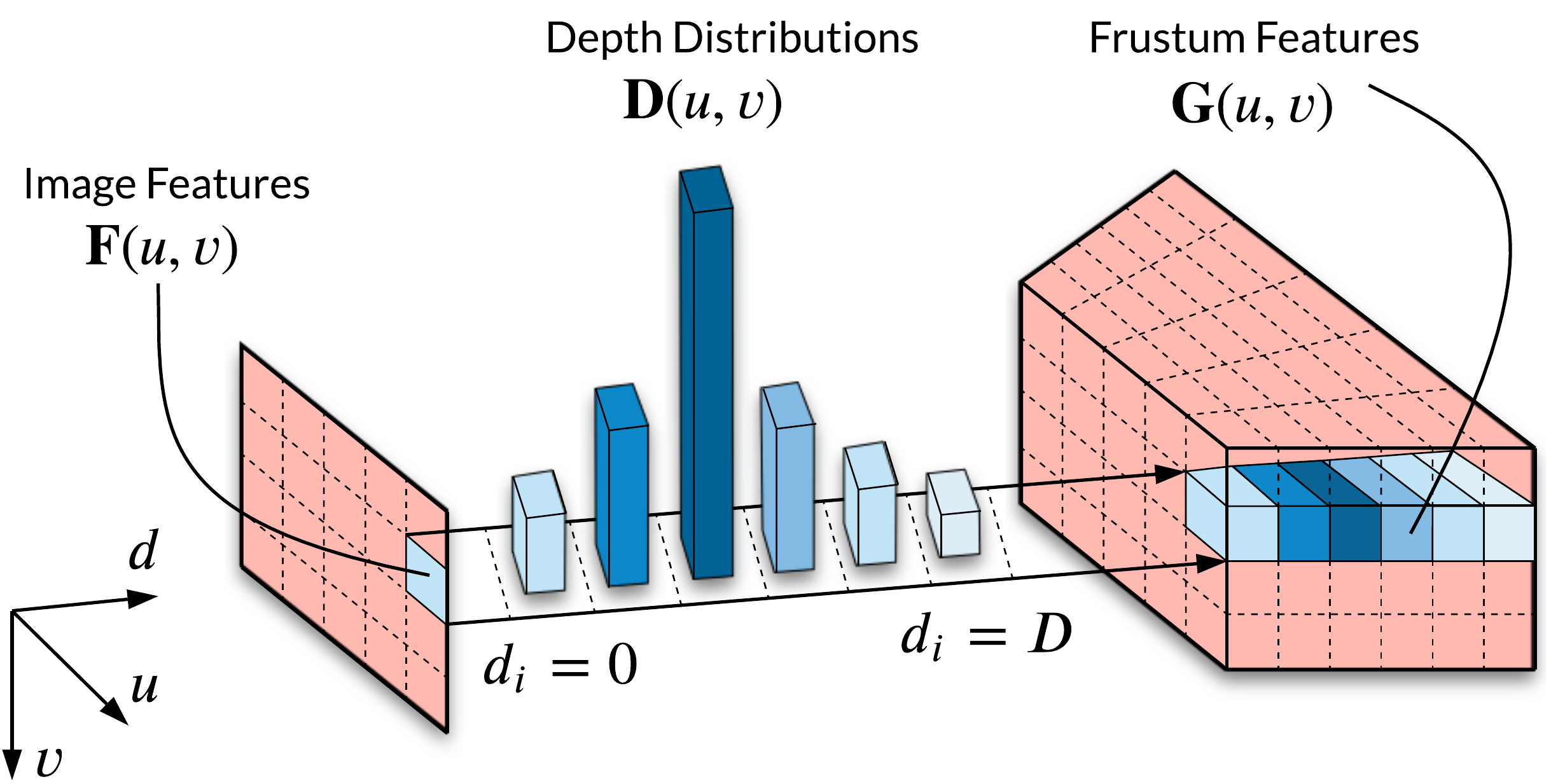}
\end{center}
\caption{An illustration of the feature lifting methods. {\it Left:} 3D features are generated by accumulating 2D features over corresponding areas. {\it Right:} image features are weighted by their depth distribution to lift the 2D features into the 3D space. From \cite{oftnet} and \cite{caddn}.}
\label{fig:feature_transoformation}
\end{figure}

\noindent
{\bf Feature lifting for monocular methods.}
\label{sec:feat_lifting_mono}
Roddick {\it et al.}~\cite{oftnet} proposed a retrieval-based detection model, named OFTNet, to achieve the feature lifting. They obtain the voxel feature  by accumulating the 2D features over the area of the front-view image feature  corresponding to the projection of each voxel's top-left corner $(u_{1}, v_{2})$ and bottom-right corner $(u_{2}, v_{2}$):
\begin{equation}
\mathbf{V}(x, y, z) =\frac{1}{(u_{2}-u_{1})(v_{2}-v_{1})}\sum^{u_{2}}_{u=u_{1}}\sum^{v_{2}}_{v=v_{1}}\mathbf{F}(u, v),
\end{equation}
where $\mathbf{V}(x, y, z)$ and $\mathbf{F}(u,v)$ denote the features for the given voxel $(x, y, z)$ and pixel $(u,v)$. Differently, Reading {\it et al.}~\cite{caddn} achieve feature lifting in a back-projection manner \cite{liftsplat}. Firstly, they discretize the continuous depth space to multiple bins and regard the depth estimation as a classification task. In this way, the output of depth estimation is the distribution $\mathbf{D}$ for these bins, instead of a single value. Then, each feature pixel $\mathbf{F}(u, v)$ is weighted by its associated depth probabilities  $\mathbf{D}(u, v)$ to generate the 3D frustum feature $\mathbf{G}(u, v)$:
\begin{equation}
\mathbf{G}(u, v)=\mathbf{D}(u, v) \otimes \mathbf{F}(u, v),
\end{equation}
where $\otimes$ denotes the outer product. Note this frustum feature is based on the image-depth coordinate system $(u, v, d)$, and it needs to be further aligned to the 3D world coordinate system $(x, y, z)$ using camera parameters to generate the voxel feature or BEV feature.  Recently, Huang {\it et al.} \cite{bevdet} adopted this lifting method and built BEV pipeline, which achieves promising performance on the multi-camera setting of image-based 3D detection. 
Figure \ref{fig:feature_transoformation} visualizes these two methods.

\noindent
{\bf Feature lifting for stereo methods.}
\label{sec:feat_lifting_stereo}
Thanks to the well-developed stereo-matching technologies, building 3D features from stereo pairs is easier to achieve than building them from monocular images. Chen {\it et al.} \cite{dsgn} proposed the Deep Stereo Geometry Network (DSGN), achieving the feature lifting with stereo images as input. They first extract feature from the stereo pairs and then build 4D plane-sweep volume following the classic plane sweeping approach \cite{spacesweep,deepstereo,mvsnet} by concatenating the left image feature and the reprojected right image feature at equally spaced depth values. Then, this 4D volume will be transformed into the 3D world space before generating the BEV map which used to predict the final results.

\subsubsection{Data Lifting-based Methods}
\label{sec:framework_data}

In the data lifting-based methods, the 2D images are transformed into the 3D data ({\it e.g.} the point cloud). Then the 3D features are extracted from the resulting data. In this section, we first introduce the pseudo-LiDAR pipeline, which lifts the images to point clouds, and the improvements designed for it. Then we introduce the image representation-based methods and other lifting schemes.

\begin{figure}
\centering 
\resizebox{\linewidth}{!}{
\includegraphics{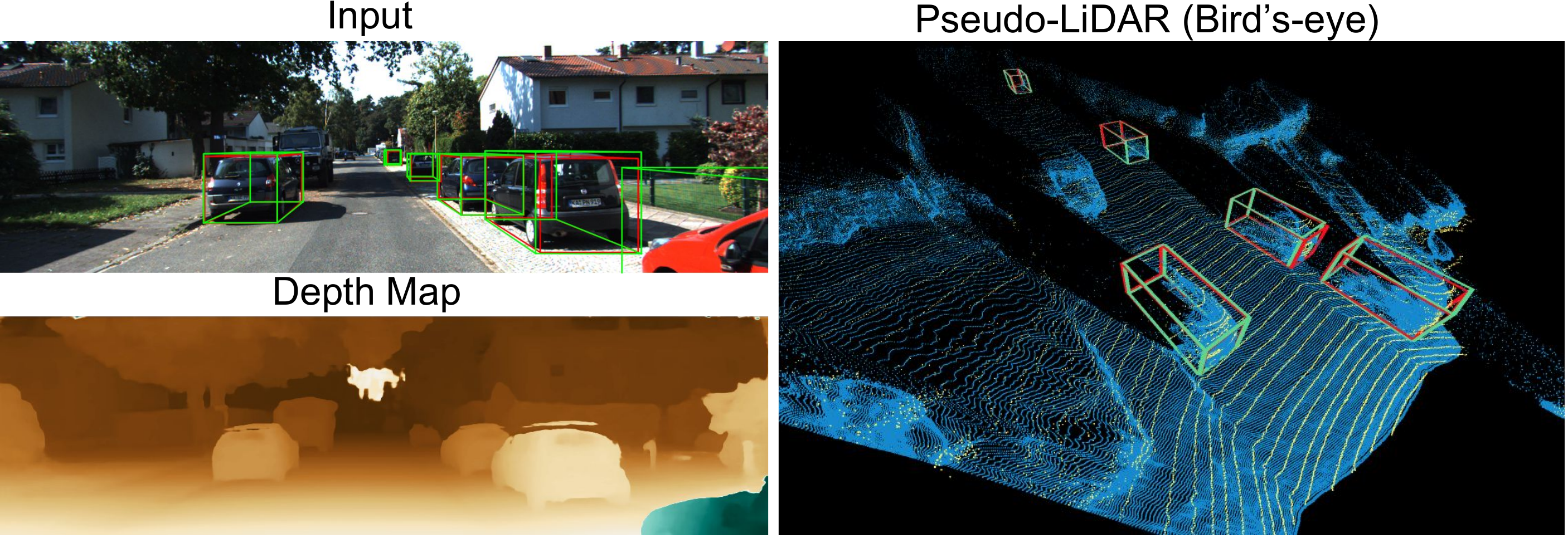}}
\caption{Comparison of different data representations: RGB image ({\it top left}), depth map ({\it bottom left}), and pseudo-LiDAR ({\it right}). From \cite{pseudolidar}.}
\label{fig:data_representation}
\vspace{-4pt}
\end{figure}

\noindent
{\bf The pseudo-LiDAR pipeline.}
\label{sec:framework_pseudolidar}
Thanks to the well-studied depth estimation, disparity estimation, and LiDAR-based 3D object detection, a new pipeline \cite{pseudolidar,am3d,monowithpl} was proposed to build a bridge between the image-based methods and LiDAR-based methods. In this pipeline, we first need to estimate the \emph{dense} depth maps~\cite{monodepth,dorn} (or disparity maps \cite{dispnet,pspnet} and then transform them into the depth maps \cite{pseudolidar}) from images. Then, the 3D location $(x, y, z)$ of the pixel $(u, v)$ can be derived using Equation \ref{equ:2dto3d}.
By back-projecting all the pixels into 3D coordinates, the {\it pseudo-LiDAR} signals $\{(x^{(n)}, y^{(n)},z^{(n)})\}_{n=1}^{N}$ can be generated, where $N$ is the number of pixels. After that, LiDAR-based detection methods \cite{frustumpointnet,avod,pixor,pointrcnn} can be applied using the pseudo-LiDAR signals as input. The comparison of data representations used in this pipeline is shown in Figure~\ref{fig:data_representation}. The success of the pseudo-LiDAR pipeline shows the importance of spatial features in this task, and breaks the barrier between images-based methods and LiDAR-based methods, which makes it possible to apply the advanced technologies of another field. See Appendix \ref{supp:dis_pseudolidar} for more discussion about image, pseudo-LiDAR, and LiDAR representations.

\noindent
{\bf Improving the quality of depth maps (or resulting pseudo-LiDAR signals).}
Theoretically, the performances of pseudo-LiDAR-based models heavily rely on the quality of depth maps and some works \cite{pseudolidar,am3d,wasserstein} had confirmed this by adopting different depth estimators. Except for the improvement of depth estimation \cite{monodepth,dorn,bts} and stereo matching \cite{disprcnn,pspnet,wasserstein}, there are some other methods that improve the quality of depth maps. Note that a small error in disparity will lead to a large error in depth for the far-away objects, which is the primary weakness of the pseudo-LiDAR-based methods. To this end, You {\it et al.} \cite{pseudolidar++} propose to transform the disparity cost volume to depth cost volume and learn depth directly end-to-end instead of through disparity transforms. Peng {\it et al.} \cite{ida3d}  use a non-uniform disparity quantization strategy to ensure a uniform depth distribution, which can also reduce the disparity-depth transformation errors for the far-away objects. Besides, directly improving the accuracy of pseudo-LiDAR signal is another option. For this, You {\it et al.} \cite{pseudolidar++} propose to use the cheap sparse LiDAR ({\it e.g.} 4-beam LiDAR) to correct the systematic bias in depth estimator. These designs can significantly boost the accuracy of generated pseudo-LiDAR signals, especially for the far-away objects.

\noindent
{\bf Focusing on the foreground objects.}
The original pseudo-LiDAR model estimates the full disparity/depth maps for the input images. This choice introduces lots of unnecessary computational cost and may distract the networks from the foreground objects, because only the pixels corresponding to the foreground objects are the focus in the subsequent steps. Based on this observation, several methods proposed their improvements. Specifically, similar to the LiDAR-based 3D detector F-PointNet \cite{frustumpointnet}, Ma {\it et al.} \cite{am3d} use the 2D bounding box to remove the background points. Besides, they also propose a scheme based on dynamic threshold to further remove the noise points. Compared with the 2D bounding box, the methods in~\cite{monowithpl,zoomnet,oc_stereo,disprcnn} adopt instance mask, which is a better filter but requires additional data with ground-truth masks (Section \ref{sec:aux_data} discusses how to use auxiliary data to generate the mask annotations)

Besides, the methods in \cite{foresee,gcstereo} propose to address this problem in the depth estimation phase. They divide the pixels of the input images into foreground and background using 2D bounding boxes as masks, and apply higher training weight for the foreground pixels. Consequently, the depth values of foreground regions are more accurate than the baseline, thereby improving the 3D detection performance.

\noindent
{\bf Aggregation with other information.}
As described before, most pseudo-LiDAR-based methods only adopt the resulting pseudo-LiDAR signals as input. Another improvement direction is to enrich the input data with other information. Ma {\it et al.} \cite{am3d} fuse the RGB features of each pixel to its corresponding 3D point using an attention-based module. Besides, a RoI-level RGB feature is also used to provide the complementary information to pseudo-LiDAR signals. Pon {\it et al.} \cite{zoomnet} propose to use the pixel-wise part location map to augment the geometric cues for the pseudo-LiDAR signals (similar idea to the LiDAR-based 3D detector \cite{points2part}). In particular, they use a CNN branch to predict the relative position of each pixel/point of the 3D bounding box, and then use this relative position to enrich the pseudo-LiDAR signals.

\noindent
{\bf End-to-end training.}
Generally, the pseudo-LiDAR-based methods are clearly divided into two separate parts: depth estimation and 3D detection, and cannot be trained end-to-end. For this problem, Qian {\it et al.} \cite{pseudolidare2e} propose a differentiable Change of Representation (CoR) module that allows the back-propagation of gradients from 3D detection network to depth estimation network, and the whole system can benefit from joint training.

\noindent
{\bf Image representation-based methods.}
To explore the underlying reasons for the success of pseudo-LiDAR-based methods, Ma {\it et al.} \cite{patchnet} proposed PatchNet, an image representation-based equivalent implementation of the original pseudo-LiDAR model \cite{pseudolidar}, and achieved the almost same performance. Based on this, Ma {\it et al.} argue that the data lifting in Equation~\ref{equ:2dto3d}, which lifts the 2D location in the image coordinate to the 3D location in the world coordinate, is the key of the success of pseudo-LiDAR family, instead of the data representation. Simonelli {\it et al.} \cite{demystifying} extend PatchNet by re-scoring the confidence of 3D bounding boxes with a confidence head, and achieve better performance. Note that most of the designs in Section~\ref{sec:framework_pseudolidar} for pseudo-LiDAR-based methods can be easily used in image representation-based method. Besides, benefiting from the well-studied 2D CNN designs, the image-based data lifting model may have greater potentials~\cite{patchnet}.

\noindent
{\bf Other lifting schemes.}
Different from previously introduced models which achieve the data lifting by depth estimation and Equation \ref{equ:2dto3d}, Srivastava {\it et al.}~\cite{birdgan} introduce another way for data lifting. Specifically, they transform the front-view images into the BEV maps using Generative Adversarial Nets (GAN)~\cite{gan,dcgan}, where the generator network aims to generate the BEV maps correspond to the given images and the discriminator network serves to classify the BEV maps are generated or not. 
Besides, Kim {\it et al.} \cite{inverse_mapping} propose to use inverse perspective mapping to transform the front-view images into BEV images.
After obtaining the BEV images, these two works can use the BEV-based 3D detectors, like MV3D~\cite{mv3d} or BirdNet~\cite{birdnet} to estimate the final results.

%% file: components/components.tex
\section{Comparison of components}
\label{sec:components}
In this section, we provide the detailed comparison for each required component of the 3D object detectors. Compared with the framework-level designs, the following designs are usually modular and can be applied to different algorithms flexibly.

\subsection{Feature Extraction}
\label{sec:components_feature}
Same as other tasks in the CV community, a good feature representation is a key factor in building high-performance image-based 3D detectors.
The majority of recent methods use standard CNNs as their feature extractors, while some methods deviated from this introducing better features extraction methods. We will briefly cover them here.

\subsubsection{Standard Backbone Nets}
Although generally the input data is only the RGB image, the feature lifting-based methods and data lifting-based methods facilitate the use of 2D CNNs \cite{resnet,densenet,resnext,dcn,dla}, 3D CNNs  \cite{3dcnn,second}, and point-wise CNNs \cite{pointnet,pointnet++,rsnet} as the backbone networks.
Among the standard backbones, DLA \cite{dla} and ResNet \cite{resnet} are generally used to extract 2D features, and the sparse 3D conv \cite{second} is the most popular backbone for 3D feature extraction. Note that the 2D backbones can also be used in the methods based on 3D features. For example, the feature-lifting-based methods, {\it e.g.} DSGN \cite{dsgn} and CaDDN \cite{caddn}, first use ResNet to extract 2D features from images, and then use 3D convolutions to generate more discriminative features after lifting the 2D features into 3D features.



\subsubsection{Local Convolution}
Brazil and Liu \cite{m3drpn} propose to use two parallel branches to extract the spatial-invariant features and spatial-aware features respectively. In particular, to better capture the spatial-aware cues from monocular images, they further propose a local convolution: the depth-aware convolution. The proposed operation uses non-shared convolution kernels to extract the features for the different rows (roughly corresponding to different depths) in the feature space. Finally, the spatial-aware features are combined with the spatial-invariant ones before estimating the final results. Note that the non-shared kernels will introduce extra computational costs, and \cite{m3drpn} also propose an efficient implementation for this scheme.

\subsubsection{Feature Attention Mechanism}
\label{sec:comp_attention}
Since Hu {\it et al.} \cite{senet} introduced the attention mechanism \cite{attention} to CNN, lots of attention blocks \cite{senet,cbam,aaconv} are proposed. Although the details of these methods are different, they usually share the same key idea: re-weighting the features along a specific dimension, {\it e.g.} channel dimension.

Qin {\it et al.} \cite{tlnet} propose an attention scheme for 3D detection from stereo pairs. In particular, they calculate the correlation score $s_{i}$ for the $i^{th}$ channel of the left-image feature $\mathbf{F}_{i}^{l}$ and the right-image feature $\mathbf{F}_{i}^{r}$ using the cosine similarity:
\begin{equation}
s_{i}=\cos <\mathbf{F}_{i}^{l}, \mathbf{F}_{i}^{r}>=\frac{\mathbf{F}_{i}^{l} \cdot \mathbf{F}_{i}^{r}}{\left\|\mathbf{F}_{i}^{l}\right\| \cdot\left\|\mathbf{F}_{i}^{r}\right\|}.
\end{equation}
Then, the features are scaled by the scaling factor $s_{i}$. Unlike other attention modules that learn the scaling factors in a data driven manner, this scheme updates the features using the correlation between left-image and right-image features, thus more interpretable. Besides, this design is also been adopted by other stereo-based 3D detection methods \cite{ida3d}.

\subsubsection{Depth Augmented Feature Learning}
\label{sec:components_feature_depth}
To provide the depth cues unavailable in the RGB images, an intuitive scheme is using the depth maps (generally obtained from an off-the-shelf model or a sub-network) to augment the RGB features \cite{multifusion,roi10d}. Besides, some efficient depth-augmented feature learning methods are proposed for this purpose. In particular, Ding {\it et al.} \cite{d4lcn} propose a local convolutional network, where they use the depth maps as guidance to learn the dynamic local convolutional filters with different dilated rates for RGB images. Wang {\it et al.} \cite{ddmp3d} design a message-passing module between RGB features and depth features based on the graph neural network (GNN). Speciﬁcally, they regard the feature vector at each position and its most-relevant neighborhoods as the nodes of GNN. After dynamically sampling the nodes from image and depth features, they use GNN to propagate the depth cues to RGB features. Finally, they apply this module in multiple feature levels and obtain richer features for 3D detection. 

\subsubsection{Feature Mimicking}
Recently, some methods propose to learn the features of image-based models under the guidance of LiDAR-based models. Particularly, Ye {\it et al.} \cite{da3ddet} adopt the pseudo-LiDAR (data lifting) pipeline and enforce that the features learned from pseudo-LiDAR signals should be similar to those learned from real LiDAR signals. Similarly, Guo {\it et al.} \cite{ligastereo} apply this mechanism to the feature lifting-based method and conduct the feature mimicking in the transformed voxel features (or BEV features). 
Furthermore, Chong {\it et al.} \cite{monodistill} generalize this scheme to the result-lifting methods.
They all transfer the learned knowledge from the LiDAR-based models to image-based models in the latent feature space, and the success of these works shows that image-based methods can benefit from feature mimicking. 

\subsubsection{Feature Alignment}
\begin{figure}
\begin{center}
\resizebox{0.99\linewidth}{!}{
\includegraphics{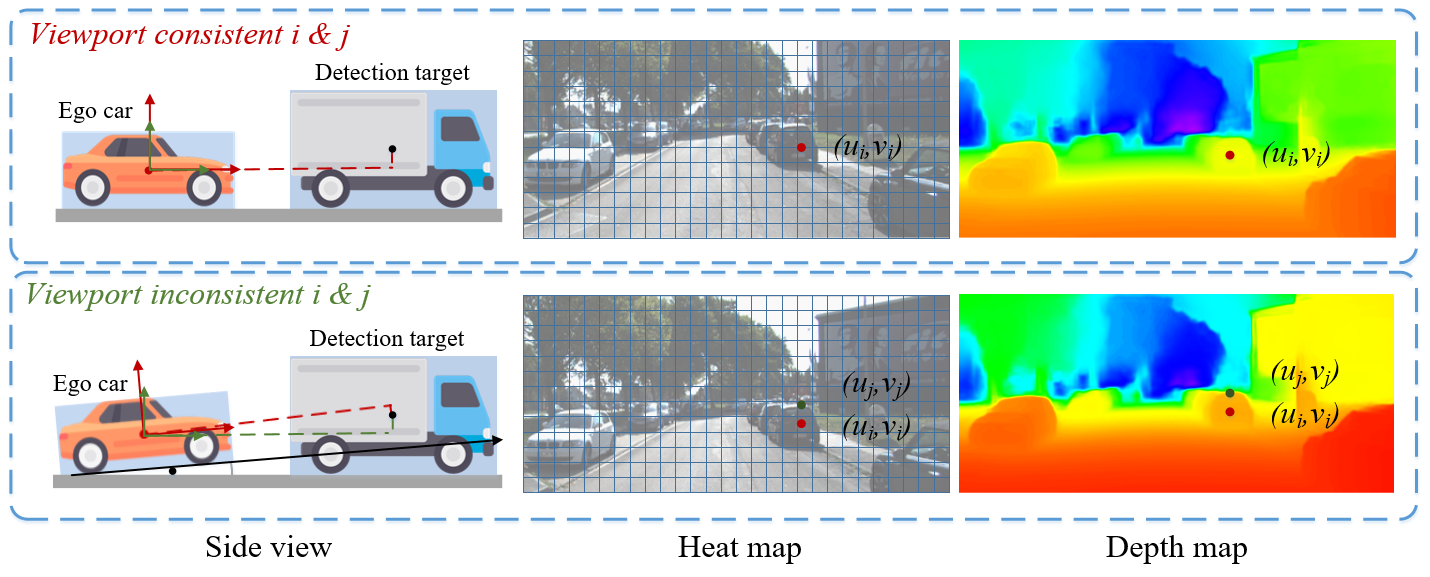}}
\end{center}
\caption{3D detection assumes the ground plane is flat and only the yaw angle is considered ({\it top}). However, in real-world applications, there are some uneven roads, which leads to a bias between the object's actual position and computed position ({\it bottom}). From \cite{monoef}}
\label{fig:monoef}
\end{figure}
As introduced in Section~\ref{sec:task}, only the yaw angle is considered in the 3D detection task. However, this design will cause a misalignment problem when the roll/pitch angle is not zero, Figure~\ref{fig:monoef}  illustrates this problem. For this problem, Zhou {\it et al.} \cite{monoef} propose a feature alignment scheme. In particular, they first estimate the ego-pose using a sub-network, and then design a feature transfer net to align the features in both content level and style level based on the estimated camera pose. Finally, they use the rectiﬁed features to estimate the 3D bounding boxes.

\subsubsection{Feature Pooling}
\begin{figure}
\begin{center}
\resizebox{0.9\linewidth}{!}{
\includegraphics{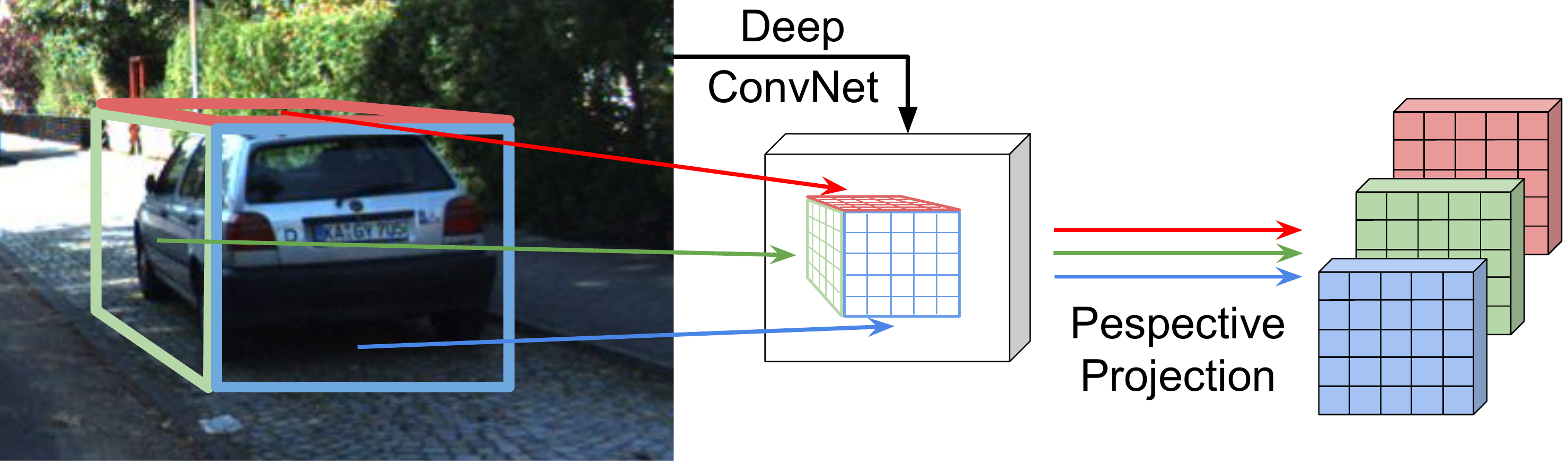}}
\end{center}
\caption{GS3D extracts features from the visible surfaces of the projected 3D bounding boxes and uses them to predict the final results. From \cite{gs3d}.}
\label{fig:gs3d}
\end{figure}
Li {\it et al.} \cite{gs3d} propose a new feature pooling scheme for image-based 3D detection. As shown in Figure \ref{fig:gs3d}, for a given 3D anchor, they extract the features from the visible surfaces and warp them into a regular shape ({\it e.g.} $7\times7$ feature map) by perspective transformation. Then, these feature maps are combined and used to refine the proposals to the final results. Note that these features can be further augmented by concatenating the features extracted from 2D anchors using RoI Pool \cite{fasterrcnn} or RoI Align \cite{maskrcnn}.

\subsection{Result Prediction}
After the CNN features are obtained, the 3D detection results are predicted from the extracted features. In this section, we group the novel designs for the result prediction into different aspects and discuss these methods in detail.

\subsubsection{Multi-Scale Prediction}
A baseline model is to predict the results using the features of the last CNN layer \cite{multifusion,pseudolidar,patchnet,demystifying}. However, a major challenge of this scheme comes from the varied scales of the objects. Particularly, CNNs commonly extract the features layer by layer, which leads to different receptive fields and semantic levels for features at different layers. Consequently, it is hard to predict all the objects using the features from a specific layer. To address this issue, lots of methods have been proposed, broadly grouped into layer-level methods and kernel-level methods.

\noindent
{\bf Layer-level methods.}
The first group of methods mainly operates on the layer-level of CNNs and can be subdivided into the following three sub-groups.

\begin{figure}
\centering 
\resizebox{\linewidth}{!}{
\includegraphics{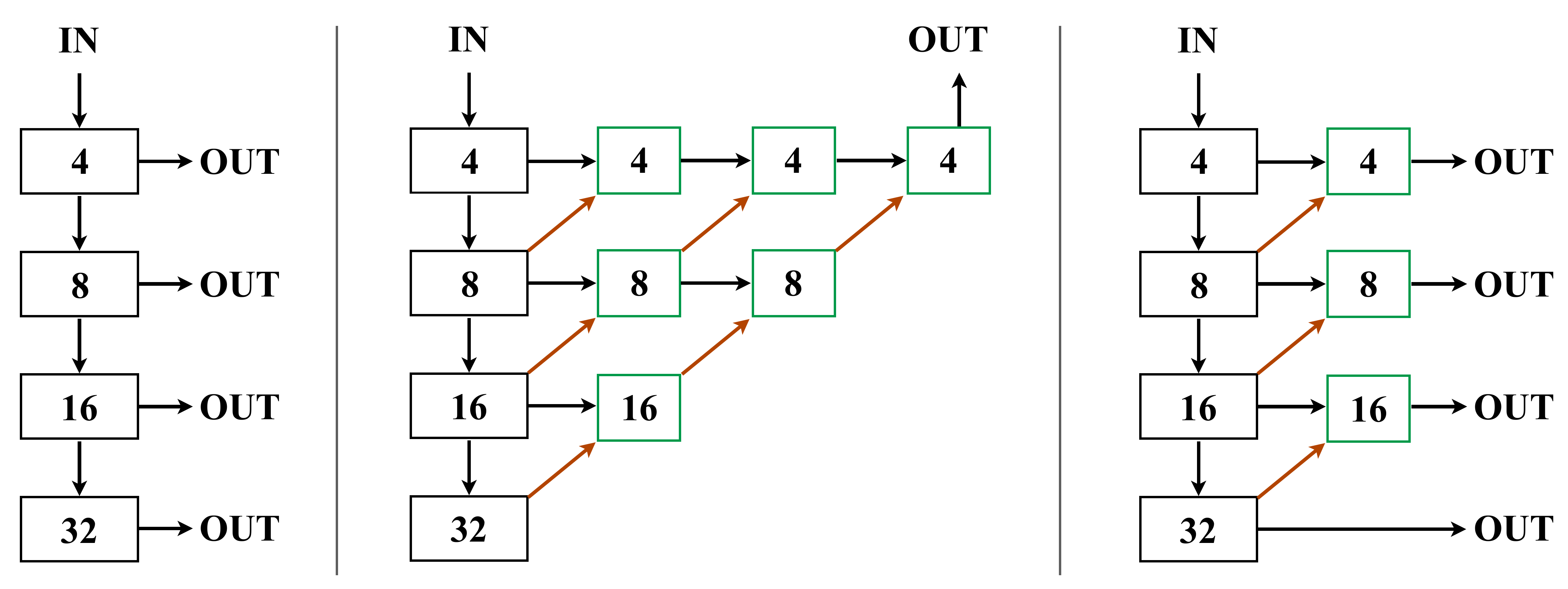}}
\caption{Illustration of the multi-level prediction ({\it left}), feature-fusion ({\it middle}), and hybrid ({\it right}) designs for multi-scale detection. The green rectangle and red arrow respectively denote the feature aggregation node and up-sampling operation.}.
\label{fig:multiscale}
\vspace{-5pt}
\end{figure}

\noindent
{\it Multi-level prediction-based models.}
Liu {\it et al.} \cite{ssd} and Cai {\it et al.} \cite{mscnn} propose to use a multi-layer prediction mechanism to address this problem, where each layer focuses on a specific range of scales. Figure \ref{fig:multiscale} ({\it left}) shows the main idea of this design.

\noindent
{\it Feature fusion-based models.}
Another popular solution is to aggregate the features from different layers and predict all samples using this augmented feature map. Figure \ref{fig:multiscale} ({\it middle}) visualizes a typical method \cite{dla} of this family. Note that lots of image-based 3D detectors \cite{oftnet,monopsr,am3d,m3drpn,monogrnet,monopair,d4lcn,smoke,monodle,monogeo,gupnet,autoshape,monoflex,m3dssd,rtm3d} adopted this method for its simple and efficient design. 

\noindent
{\it Hybrid models.}
In fact, recent approaches rarely use only one strategy, and the hybrid models are more welcomed. For example, FPN \cite{fpn}, shown in Figure \ref{fig:multiscale} ({\it right}), combines the multi-level prediction and feature fusion scheme, and the FPN-like schemes are embraced by several 3D detection models \cite{monodis,monodis_journal,ida3d,roi10d,stereorcnn,fcos3d,dd3d}.

\noindent
{\bf Kernel-level methods.}
Some methods try to solve this problem by adjusting the receptive field at kernel-level. The dilated convolution \cite{deeplabv2} (also called `atrous' convolution) is a pioneering work which is initially proposed to extract multi-scale features for semantic segmentation task. It introduces another parameter, the dilation rate, which controls the sampling interval in the convolution operation. This design can enlarge the receptive ﬁeld without introducing extra computational cost. Ding {\it et al.} \cite{d4lcn} introduce this convolution to monocular 3D detection and propose a scheme to dynamically adjust the dilated rate for each object according to its depth value. Besides, Dai {\it et al.} \cite{dcn,dcnv2} propose the deformable convolution, which allows the convolution kernels to learn their sampling positions in a data driven manner. Luo {\it et al.} \cite{m3dssd} propose a variant of deformable convolution, which generates the sampling positions according to the shape of the anchors. Note that the dilated convolution can be regarded as a static special case of the deformable convolution. Besides, the kernel-level designs are orthogonal to the layer-level designs, which means they can collaboratively work in the same algorithm.

\subsubsection{Multi-Camera Prediction}
To cover all objects in 360$^{\circ}$ view, the multi-camera detection is adopted by recent large-scale benchmarks \cite{nuscenes,waymo}. 
A simple baseline \cite{centernet} is to treat all views as separate images and predict results from them separately. Then a global Non-Maximum Suppression (NMS) operation is applied to merge the results from different views and remove the duplicate objects. 
Recently, a BEV solution \cite{detr3d,bevdet} is proposed for the multi-camera setting. In particular, this pipeline first lifts the features of different views from the image space to the BEV space and then integrates these BEV maps into a single feature map for the whole scene. After that, the results for all views can be predicted from this integrated BEV feature map. In addition to solving the task of multi-camera detection, this scheme generally generates more discriminative features due to the self-calibration of the features among different views, and similar strategies may be applied to other scenarios in future research, {\it e.g.} 3D detection from temporal sequences or multi-modality data.

\subsubsection{Out-of-Distribution Samples}
Due to the range, truncation, occlusion, etc., different objects tend to have different characteristics, and predicting all objects from a unified network may not be optimal. Based on this problem,  \cite{patchnet,demystifying,monoflex} adopted the self-ensembling strategy. In particular, Ma {\it et al.} \cite{patchnet} divide the objects into three clusters by their depth values (or the `difficulty' levels defined by KITTI 3D dataset), and use different heads to predict them in parallel. Simonelli {\it et al.} \cite{demystifying} extended this design by adding a re-scoring module for each head. Zhang {\it et al.} \cite{monoflex} decouple the objects into two cases according to their truncation levels and apply different label assignment strategies and loss functions to them.

Besides, Ma {\it et al.} \cite{monodle} observe that some far-away objects are almost impossible to localize accurately and reducing their training weights (or directly removing these samples from the training set) can improve the overall performance. The underlying mechanism of this strategy has the same goal as \cite{patchnet,demystifying,monoflex}, {\it i.e.} to avoid the distraction from out-of-distribution samples to the model training. 

\subsubsection{Projective Modeling for Depth Estimation}
Compared with a stand-alone depth estimation task, depth estimation in 3D detection has more geometric priors, and the projective modeling is the most commonly used one. In particular, the geometric relationship between the height of 3D bounding box $H_{3D}$ and the height of its 2D projection $H_{2D}$ can be formulated as:
\begin{equation}
\label{equ:depth_from_geo}
d = f \times \frac{H_{3D}}{H_{2D}}
\end{equation}
where $d$ and $f$ respectively denote the depth of the object and the focal length of the camera. The height of the 2D bounding box is used to approximate $H_{2D}$ in \cite{decoupled3d,monopsr,ur3d,centervoting}, so they can compute a rough depth using estimated parameters. However, when the height of the 2D bounding box (denoted as $H_{bbox2D}$) is used as the $H_{2D}$ in Equation \ref{equ:depth_from_geo},  extra noise is introduced, because $H_{2D}\neq H_{bbox2D}$. To alleviate this problem, Lu {\it et al.} \cite{gupnet} propose an uncertainty-based scheme, which models the geometric uncertainty in the projective modeling. Besides, Barabanau {\it et al.} \cite{keypointsgeo} annotate the key-points of cars with the help of CAD models, and use the height difference of 2D/3D key-points to get the depth. Differently, Zhang {\it et al.} \cite{monogeo} revise Equation~\ref{equ:depth_from_geo} by considering the interaction of the locations, dimension and orientations of the objects, and build the relationship between the 3D bounding box and its 2D projection.

In brief, GUPNet~\cite{gupnet} captures the uncertainty in the noisy perspective projection modeling, Barabanau {\it et al.}~\cite{keypointsgeo} eliminate the noise by re-labeling, and Zhang {\it et al.}~\cite{monogeo} solve the error by mathematical modeling.

\subsubsection{Multi-Task Prediction}

\noindent
{\bf 3D detection as multi-task learning.}
3D detection can be seen as a multi-task learning problem because it needs to output the class label, location, dimension, and orientation together. Lu {\it et al.} \cite{gupnet} propose to dynamically adjust the learning weights of each task for the balanced learning. Different from other weight-based multi-task learning methods~\cite{gradnorm,multitask_uncert}, which assume that each task is independent from each other, there are some dependencies among the sub-tasks in 3D detection, {\it e.g.,} the height of 2D/3D bounding box can provide hints for depth estimation. Therefore, they build the hierarchical relationship of all tasks, and the training weight of each task is scheduled by its pre-tasks. Besides, Zou {\it et al.} \cite{dfrnet} divide all tasks into the appearance-specific tasks and the localization tasks, and the features for these two groups are learned separately with a message passing module.

\noindent
{\bf Joint training with other tasks.}
\label{sec:joint_training}
A number of works \cite{padnet,multitasksurvey,taskonomy} had shown that the CNN can benefit from joint training with multiple tasks. Similarly, Ma {\it et al.} \cite{monodle} observe the 2D detection can serve as an auxiliary task to monocular 3D detection and provide additional geometric cues to the neural network. Besides, Guo {\it et al.} \cite{ligastereo} find this is also effective for stereo 3D detection. Note that the 2D detection is a required component in some methods \cite{m3drpn,monodis,pseudolidar,am3d,patchnet}, instead of an auxiliary task. Based on this, Liu {\it et al.} \cite{monocon} find extra key-points estimation task can further enrich the CNN features, and the estimated key-points can be used to further optimize the depth estimation sub-task \cite{shiftrcnn,autoshape,rtm3d}.  Besides, depth estimation can also provide valuable cues to the 3D detection model. In particular, some works \cite{dd3d,dsgn,ligastereo,monodistill} conduct an extra depth estimation task to guide the shared CNN features to learn the spatial features, and Park {\it et al.} \cite{isplneeded} show that pre-training on the large-scale depth estimation dataset can significantly boost the performance of their 3D detector.

\subsection{Loss Formulation}
Loss function is an indispensable part of the data driven models, and the loss formulation of 3D detection can be simplified to :
\begin{equation}
\mathcal{L} = \mathcal{L}_{{\bf cls}} + \mathcal{L}_{{\bf loc}} +
\mathcal{L}_{{\bf dim}} +
\mathcal{L}_{{\bf ori}} + 
\mathcal{L}_{{\bf joi}} + 
\mathcal{L}_{{\bf conf}} + 
\mathcal{L}_{{\bf aux}}. \label{Eq:Loss}
\end{equation}
Particularly, the classification loss $\mathcal{L}_{{\bf cls}}$ serves to identify the category of a candidate and give the confidence. The location loss $\mathcal{L}_{{\bf loc}}$, dimension loss $\mathcal{L}_{{\bf dim}}$, and orientation loss $\mathcal{L}_{{\bf ori}}$ are designed to regress the components of the parameterization of 3D bounding box, {\it i.e.} location, dimension, and orientation respectively.  The last three loss items are optional. In particular, the loss $\mathcal{L}_{{\bf joi}}$, {\it e.g.}  corner loss \cite{roi10d}, can jointly optimize the location, dimension, and orientation in a single loss function. The confidence loss $\mathcal{L}_{{\bf con}}$ is designed to give a better confidence to the detected boxes. Finally, the auxiliary loss $\mathcal{L}_{{\bf aux}}$ can introduce additional geometric cues to CNNs. These losses in Equation \ref{Eq:Loss} are discussed below.

\subsubsection{Classification Loss $\mathcal{L}_{{\bf cls}}$}
For classification, FocalLoss \cite{focalloss} or its variant \cite{cornernet} is used by most methods. Compared with the standard cross-entropy loss, this loss function reduces the penalty on the easy cases and focuses more on the hard, misclassified examples. In this way, this loss boosts the classification accuracy.

\subsubsection{Location Loss $\mathcal{L}_{{\bf loc}}$}
The feature/data lifting-based methods generally regress the locations using L1 loss (or smooth L1 loss, L2 loss, etc. and we omit them in the following part for brevity):
\begin{equation}
\mathcal{L}_{\mathbf{loc}} = \sum_{i\in\{x, y, z\}}|| \mathbf{loc}_{i} -  \mathbf{loc}_{i}^{*}||_{1},
\label{equ:loc_loss}
\end{equation}
where $|| \cdot ||_{1}$ denotes the L1 norm. $\mathbf{loc}_{i}$ and $\mathbf{loc}_{i}^{*}$ are the estimated location and corresponding ground truth, respectively. Generally, the models predict the relative offset to a specific anchor, instead of the absolute position. As for the result lifting-based methods, the 3D location is derived from the 2D location and depth (note most of the feature lifting-based and data lifting-based methods also need depth for their transformations), and the loss function can be formulated as:
\begin{equation}
\mathcal{L}_{{\bf loc}} = \mathcal{L}_{{\bf loc_{2d}}} + \mathcal{L}_{{\bf depth}},
\end{equation}
where $\mathcal{L}_{{\bf loc_{2d}}}$ is the 2D location loss and generally shares the similar formulation as Equation \ref{equ:loc_loss}. The $\mathcal{L}_{{\bf depth}}$ is the depth loss. Since some works \cite{monodle,pgd,isplneeded} point out that depth is the key in image-based 3D detection, we mainly review the novel loss formations of this item in image-based 3D detection approaches.

\noindent
{\bf Uncertainty modeling.}
Following~\cite{uncertainties,geometry}, some works \cite{monopair,monodle,monoloco,ur3d,gupnet} model the heteroscedastic aleatoric uncertainty in depth estimation. Specifically, to capture the uncertainty, detectors should simultaneously predict the depth $\mathbf{d}$ and the standard deviation $\mathbf{\sigma}$  (or variance $\mathbf{\sigma}^{2}$):
\begin{equation}
    [\mathbf{d}, \mathbf{\sigma}] = f^{\mathbf{w}}(\mathbf{x}),
\end{equation}
where $\mathbf{x}$ is the input data, $f$ is a neural network parametrised by the parameters $\mathbf{w}$. Then, the Laplace likelihood is fixed to model the uncertainty, and the loss for the depth estimation sub-task can be formulated by: 
\begin{equation}
\mathcal{L}_{{\bf depth}} = \frac{\sqrt{2}}{\mathbf{\sigma}}||\mathbf{d} - \mathbf{d}^{*}||_{1} + \log\mathbf{\sigma},
\label{equ:laplacian}
\end{equation}
where $|| \cdot ||_{1}$ denotes the L1 norm, and $\mathbf{d}^{*}$ is the ground-truth depth. Similarly for the Gaussian likelihood \cite{monodle,gaussianyolo}:
\begin{equation}
\mathcal{L}_{{\bf depth}} = \frac{1}{2\mathbf{\sigma}^{2}}||\mathbf{d} - \mathbf{d}^{*}||_{2} + \frac{1}{2}\log\mathbf{\sigma}^{2},
\label{equ:gaussian}
\end{equation}
where $|| \cdot ||_{2}$ denotes the L2 norm (the derivation details of Equation~\ref{equ:laplacian} and Equation~\ref{equ:gaussian} can be found in \cite{geometry}, page 37). Note that this loss formulation can be applied to any regression task in theory \cite{monopair,gaussianyolo,softernms}. Besides, the uncertainty has been further utilized in other aspects, such as confidence normalization \cite{gupnet} or post-processing \cite{ur3d}.

\noindent
{\bf Discretization.}
For monocular depth estimation, Fu {\it et al.} propose DORN~\cite{dorn}, which discretizes the continuous depth values into multiple bins and considers the depth estimation as a \emph{ordinal regression} task. This model is often used as a sub-network to provide depth cues for 3D detectors. Besides, the methods in \cite{caddn,pgd} also adopt the discretization strategy, while only regarding depth estimation as a \emph{classification} task. Note that the discretization-based methods usually output the distribution of depth, instead of a single value, which can be used in feature lifting.

\subsubsection{Dimension Loss $\mathcal{L}_{{\bf dim}}$}
A common choice for the dimension loss in 3D detection is L1 loss:
\begin{equation}
\mathcal{L}_{{\bf dim}} = \sum_{i \in \{h, w, l\}} ||\mathbf{dim}_{i} - \mathbf{dim}_{i}^{*} ||_{1},
\end{equation}
where $\mathbf{dim}_{i}$ denotes the predicted dimension, and $\mathbf{dim}_{i}^{*}$ is the corresponding ground truth. The incremental method in \cite{m3drpn} compute the mean shape $[H, W, L]$ of each category first, and then estimate the residual offset of these anchors. Furthermore, Simonelli {\it et al} \cite{monodis} represent the dimension as $[He^{\delta_{h}}, We^{\delta_{w}}, Le^{\delta_{l}}]$, where $[\delta_{h}, \delta_{w}, \delta_{l}]$ are the outputs of the CNN for the objects' dimension. In this way, they can embed the physical prior, {\it i.e.} the mean shape of the objects, in the prediction and optimize the CNN's parameters in the exponential space.

Ma {\it et al.} \cite{monodle} show that the errors of different elements of the estimated dimension ({\it e.g.} height, weight, length) have different contribution rates to the change of IoU. Based on this observation, they dynamically adjust the weight of each term in this loss function, according to its partial derivative w.r.t. the 3D IoU. Besides, they also keep the absolute value of this loss function unchanged to the original L1 loss, which means that their proposed loss is the re-distribution of the standard L1 loss.

\subsubsection{Orientation Loss $\mathcal{L}_{{\bf ori}}$}
\begin{figure}
\centering
\resizebox{\linewidth}{!}{
\includegraphics{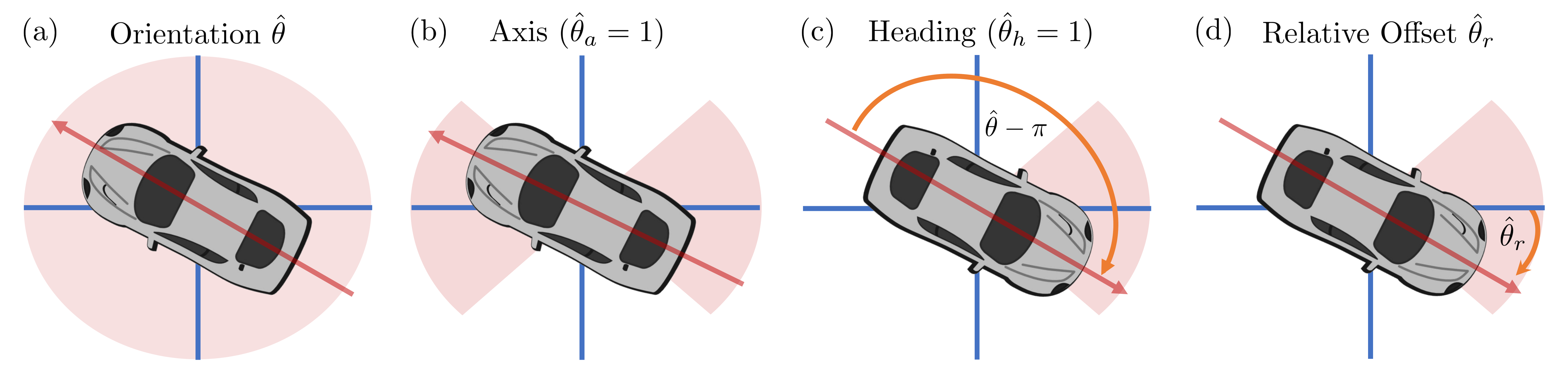}}
\caption{Illustration of the orientation formulation proposed by Brazil {\it et al.} They first classify the orientation (\emph{a}) as closer to the horizontal axis or the vertical axis (\emph{b}), and then judge whether it points in the positive or the negative direction (\emph{c}). Finally, they regress an offset to the center of the angle bin (\emph{d}). From~\cite{kinematic3d}.}
\label{fig:kin3d_heading}
\end{figure}

Compared with directly regressing the orientation of objects, the hybrid-style (classification and regression) loss formulation is the mainstream in orientation estimation, and the main difference of these losses lies in how to divide the continuous orientation into different bins \cite{sok}. In particular, Mousavian {\it et al.}~\cite{deep3dbox} divide the heading angle into $n$ overlapping bins ($n$=2 by default), and Qi {\it et al.} \cite{frustumpointnet} choose a denser, non-overlapping quantization ($n$=12 in default). These two methods are widely used in the existing 3D detectors, {\it e.g.} ~\cite{multifusion,monodle,gupnet,monopair,am3d,patchnet,pseudolidar,pseudolidar++,monowithpl,monopsr}. Besides, as shown in Figure \ref{fig:kin3d_heading}, Brazil {\it et al.} \cite{kinematic3d} propose to divide the orientation into 4 bins and use two classifiers, {\it i.e.} axis classifier (horizontal or vertical) and heading classifier (positive or negative), to find the angle interval. As for the regression part, an alternative to regressing the residual angle $\Delta \theta$ directly is to regress it in the sine and cosine spaces, {\it i.e.} $\rm{sin}(\Delta \theta)$ and $\rm{cos}(\Delta \theta)$, which has been adopted by several works, such as \cite{deep3dbox,centernet,monorcnn}. Besides, Li {\it et al.} \cite{egonet} show that orientation estimation can benefit from specific intermediate representation, {\it i.e.} interpolated cuboid.

\subsubsection{Joint Loss $\mathcal{L}_{{\bf joi}}$}
In this section, we introduce how to jointly optimize the location, dimension, and orientation in a single loss function.  

\noindent
{\bf Corner loss.}
To jointly optimize the location, dimension, and orientation, Manhardt {\it et al.} \cite{roi10d} recover the 3D coordinates of eight corners using the estimated items, and compute the corner loss:
\begin{equation}
\mathcal{L}_{{\bf corner}} = \sum^{8}_{k=1}||P_{k} - P_{k}^{*}||_{1},
\label{equ:corner}
\end{equation}
where $P_{k}$ denotes the $k^{th}$ corner of the 3D bounding box. Note that the corner loss can also be used as auxiliary loss and work together with the standard loss formulation \cite{frustumpointnet,am3d,pseudolidar,patchnet,monowithpl,da3ddet,iafa}.

\noindent
{\bf Disentangled corner loss.}
To avoid the complicated interactions between each item, Simonelli {\it et al.} \cite{monodis} propose a disentangling transformation, and this method has been adopted by some other works \cite{monodis_journal,movi3d,smoke,dd3d,demystifying}. In particular, when computing the corners, they only use one estimated item and adopt ground truths for the remaining items ({\it e.g.} using the predicted dimension and ground-truth location/orientation to compute the corners). They replicate this process for three times to separately back propagate the losses of the location, dimension, and orientation. This design removes the interactions of different items but also keeps the optimization space the same as the corner loss. Also note that this transformation can apply to any metric involving  multiple items, such as IoU.

\subsubsection{Confidence Loss $\mathcal{L}_{{\bf conf}}$}
A simple baseline method for the confidence estimation is directly adopting the classification confidence as the final score. This strategy is popular in the 2D detection field and also commonly adopted by the image-based 3D detection models, such as \cite{m3drpn,monopair,monodle}. Besides, some confidence estimation methods designed for 2D detectors are also introduced to image-based 3D detection task ({\it e.g.} following FCOS \cite{fcos}, FCOS3D \cite{fcos3d} estimates the `centerness' for each object and use it to normalize the confidence). However, these methods are generally better at representing the quality of the 2D bounding box, instead of that of the 3D bounding box. Here we represent these confidences as the 2D confidence (in fact the 2D confidence $p_{2D}$ is often adopted as the 3D confidence $p_{3D}$ directly in most works).

To better capture the quality of the estimated 3D bounding boxes, some novel designed are proposed. In particular, Simonelli {\it et al.} \cite{monodis} propose to estimate the 3D confidence given a 2D proposal $p_{3D|2D}$, and its ground truth $p_{3D|2D}^{*}$ is generated by:
\begin{equation}
    p_{3D|2D}^{*} = e^{-\frac{1}{T}\mathcal{
    L}(B, B^{*})},
\label{equ:abs_3d_conf}
\end{equation}
where T is the temperature parameter, and $\mathcal{L}(B, B^{*})$ is the disentangled corner loss of bounding box $B$ and its ground-truth $B^*$. Then the cross-entropy loss is used to optimize the CNNs, and the final 3D confidence is:
\begin{equation}
p_{3D} = p_{2D} \cdot  p_{3D|2D}.
\end{equation}
Furthermore, Simonelli {\it et al.} \cite{demystifying} use the normalized ranking of $\mathcal{L}(B, B^{*})$ as $p_{3D|2D}^{*}$, and report that this relative 3D confidence performs better than the absolute version. Besides, since the depth estimation is the bottleneck of the image-based 3D detection \cite{monodle,pgd,am3d,pseudolidar,multifusion}, Lu {\it et al.} \cite{gupnet} use the depth confidence $p_{depth}$ to replace $p_{3D|2D}$. In particular, they capture the depth uncertainty $\sigma$ using Equation \ref{equ:laplacian} and use the normalized uncertainty $e^{-\sigma}$ as the depth confidence, and the final 3D confidence is computed as:
\begin{equation}
p_{3D} = p_{2D} \cdot  p_{depth}.
\end{equation}

\subsubsection{Auxiliary Loss $\mathcal{L}_{{\bf aux}}$}

\noindent
{\bf Dense depth loss.}
Although the \emph{dense} depth estimation is unnecessary in the design of most image-based 3D detectors, several works observe that it benefits the detectors in performance. In particular, some feature lifting-based methods \cite{dsgn,caddn} find that applying the dense depth supervision is helpful to align the 2D space and the 3D space for the feature lifting models. Park {\it et al.} \cite{dd3d} find that dense depth estimation can serve as an effective pre-training task for monocular 3D detection. Besides, \cite{monodistill} reports that a separate prediction head supervised by the dense depth maps can effectively introduce the spatial cues to the CNNs, thereby improving the performance.

\noindent
{\bf 2D/3D consistency loss.}
Based on the geometric prior that the projection of the 3D bounding box should tightly fit the 2D bounding box, we can build an auxiliary loss by comparing the consistency of them. Weng and Kitani \cite{monowithpl} apply this loss function in their 3D detector, and Brazil and Liu \cite{m3drpn} use this loss in their post-processing method.

\noindent
{\bf Others.}
As introduced in Section \ref{sec:joint_training}, some works report that joint training with some other tasks, such as 2D detection\cite{monodle,dsgn} and key-points estimation\cite{monocon}, can boost the performance of 3D detection. From the perspective of loss function, the losses of these tasks can be regarded as the auxiliary losses of 3D detection.

\subsection{Post-processing}
After getting the results from CNNs, some post-processing steps are applied to remove redundant detection results or refine detection results. These steps can broadly be divided into two groups:  Non-Maximum Suppression (NMS) and post-optimization.
We present the original NMS algorithm and its variants in Appendix \ref{supp:nms} and mainly summarize the post-optimization methods in this section.

\subsubsection{Post-optimization}
\label{sec:components_postproc}
To boost the quality of detected boxes, some methods choose to further refine the outputs of CNNs by building geometric constraints in the post-optimization step.

Brazil and Liu \cite{m3drpn} proposed a post-optimization method to tune the orientation $\theta$ based on the consistency between the projected 3D bounding box and 2D bounding box. In particular, they iteratively add a small offset to the predicted orientation $\theta$ and project the updated 3D boxes into the 2D image plane. Then, they choose to accept this update or adjust the offset by checking whether the similarity between the 2D bounding box and the projected 3D bounding box increases or decreases.

Another post-optimization method is built on the one-to-one matching of the key-points of objects in the 2D image plane and 3D world space.
Specifically, in addition to the 3D boxes, Li {\it et al.} \cite{rtm3d} also estimates the projected corners in the 2D image space. After that, they project the 3D boxes into the image plane and update the estimated parameters by minimizing the pixel distances of the paired pixels, {\it i.e.} 2D/3D corners, using Gauss-Newton \cite{gn} or Levenberg-Marquardt algorithm \cite{lm}.

Chen {\it et al.} \cite{monopair} propose an object-level pair-wise constraint for their post-optimization. In particular, they regard two adjacent objects as an object pair, and additionally estimate the midpoints of the paired objects in their CNN model. After that, they can fine-tune the locations by aligning the paired objects and their midpoint. Further, they also model the uncertainties of location-related items (depth and the center of the 2D projection in \cite{monopair}) using Equation \ref{equ:laplacian}, and use the captured uncertainties as weights of the objectives in their post-optimization method. The post-optimization methods in~\cite{rtm3d,monopair} can be efficiently implemented using the open-sourced toolbox g2o \cite{g2o}.


%% file: components/aux_data.tex
\section{Leveraging Auxiliary Data}
\label{sec:aux_data}

3D object detection from RGB images is a challenging task due to the lack of  the depth information in the input data. To estimate the 3D bounding boxes more accurately, lots of methods have tried to apply auxiliary data and extract complementary features from RGB images. This section summarizes the different uses of auxiliary data and reviews the 3D detection methods according to the types of auxiliary data considered.

\subsection{LiDAR Signals}
\label{sec:auxdata_lidar}
LiDAR data is rich in spatial cues, which is missing in image data. Although LiDAR signals are unavailable for image-based methods in real application scenarios, we still can leverage them in \emph{training phase}. Overall, the related technical contributions can be roughly divided into the following categories.

\noindent
{\bf Generating depth annotations from LiDAR signals.}
As introduced in Section~\ref{sec:frameworks}, the depth maps are required by almost all methods, and a sub-task of depth estimation is always needed. Generally, the ground truths of depth values can be derived from the locations of the 3D objects. However, the ground-truth depth maps generated in this way are very sparse (less than 10 valid pixels  on average for a standard KITTI 3D image). This makes extremely difficult to train an accurate and stable depth estimator. To alleviate this problem, several methods~\cite{dsgn,caddn,pseudolidar,pseudolidar++,pseudolidare2e,ligastereo,monodistill} project the LiDAR signals to the image plane, and get the depth (or disparity) values of corresponding pixels. In this way, for KITTI 3D dataset, about additional 30K valid pixels (per image) with depth annotation can be generated. Furthermore, depth completion algorithms \cite{ip_basic} can be used to generate the dense depth map with the projected LiDAR signals as input.

\noindent
{\bf Generating mask annotations from LiDAR signals.}
Some works \cite{frustumpointnet,zoomnet,oc_stereo,pseudolidar} use the LiDAR signals to generate the instance masks. In particular, the semantic labels of LiDAR points can be determined by checking whether they are located in the 3D bounding boxes or not. Then they project the points into the image plane and assign the mask labels to their corresponding image pixels.

\noindent
{\bf Providing additional guidance in the training phase.}
The models trained from LiDAR signals can serve as the teacher network of the image-based models, and transfer the learned spatial cues to the image-based methods with the knowledge distillation mechanism \cite{kd}. Specifically, Ye {\it et al.} \cite{da3ddet} apply this strategy in the pseudo-LiDAR pipeline by mimicking the features learns from the real LiDAR. Guo {\it et al.} \cite{ligastereo} build the 3D voxel features from stereo images with the design introduced in Section \ref{sec:feat_lifting_stereo}, and then use the voxel-based LiDAR 3D detector \cite{second} to provide additional guidance in the feature level. Theoretically, this design can also be applied to other feature lifting-based models \cite{oftnet,caddn,bevdet}. Besides, MonoDistill \cite{monodistill} proposes another scheme to align the feature representations of image-based models and LiDAR-based models. This method chooses to project the LiDAR points into the image plane, and then use the identical network with the image-based detector to process the resulting data. After that, the knowledge distillation can be applied to transfer the valuable cues from the LiDAR models to image-based models, including the feature level and result level.

\noindent
{\bf Achieving data lifting with GAN.}
Except for the methods introduced in  \ref{sec:framework_data}, GAN provides a potential choice for data lifting \cite{birdgan}. In particular, the generator network aims to output the 3D representation of the input 2D images, and the discriminator network is responsible for identifying whether the data is generated or not by comparing it with real LiDAR signals.  In this pipeline, LiDAR signals are used as the input of the discriminator network in the training phase and will be removed in the inference phase. The output of the generator network will be fed to an off-the-shelf LiDAR-based 3D detector to obtain the final results.




\subsection{Temporal Sequences}

Temporal cues are vital for human visual system, and a recent work \cite{kinematic3d} has applied temporal sequences to monocular 3D detection. Particularly, Brazil {\it et al.} \cite{kinematic3d} first use the modified M3D-RPN \cite{m3drpn} to estimate the 3D boxes from separate images, and then concatenate features of adjacent frames and estimate the ego-motion of the camera. Finally, they use the 3D Kalman filter \cite{kalman} to update the estimated boxes, considering the motions of objects and ego-motion together.

Note that the initial results are still predicted from single frames, and the 3D Kalman filter is mainly used to update the results for the temporal consistency of results between image sequences. From this perspective, this method can be regarded as a kind of post-processing. Besides, benefiting from the 3D Kalman filter, this method can also predict the velocity of objects without any annotations.

\subsection{Stereo Pairs}
To provide the depth cues lacking in the monocular images, some methods adopt stereo pairs in their models. The main applications of them can be grouped into two categories: generating better depth maps and enriching the features.

\subsubsection{Generating Better Depth Maps from Stereo Pairs}
As mentioned before, many methods, including the pseudo-LiDAR-based methods (Section \ref{sec:framework_pseudolidar}) and depth-augmented methods (Section \ref{sec:components_feature_depth}), require depth maps as input. Compared with the monocular depth estimation, the depth maps estimated from stereo images \cite{dispnet,pspnet,wasserstein} are generally more accurate, and the experiments in several works \cite{multifusion,am3d,pseudolidar} demonstrate that high-quality depth maps (especially for the foreground regions \cite{foresee}) can significantly boost the final performances of these methods.

\subsubsection{Enriching the Feature Maps using Stereo Pairs}
Enriching the features maps is another application direction of the stereo images. In this family, lots of methods propose their solutions in different ways, including but not limited to feature fusion \cite{stereorcnn,dispnet}, attention mechanism \cite{tlnet,ida3d} (see Section \ref{sec:comp_attention}), and building better feature representation (such as cost volume \cite{ida3d,dsgn,ligastereo}).

\subsection{Pre-Training Data}
\label{sec:pretraining}
It is well known that pre-training from larger datasets usually improves the representation ability of neural networks and alleviates the overfitting problem. Based on this, some works choose to pre-train their models from external datasets. 
This section will summarize the main applications of pre-training in the image-based 3D detection field.

\subsubsection{Fixed Pre-trained Sub-Models}
To provide extra features, some works embed a pre-trained model in their detectors. Here we introduce the common choices for existing methods.

\noindent
{\bf Depth estimator.}
Depth estimation is a core sub-problem for image-based 3D object detection, and lots of works have shown that introducing a pre-trained depth estimator can significantly improve the accuracy of 3D detectors. In particular, some work pre-train a standalone CNN \cite{monodepth, dorn, bts} on the KITTI Depth, which contains 23,488 training samples with dense annotation for depth estimation, and use the pre-trained depth estimator to generate depth maps for KITTI 3D. After that, These estimated depth maps can serve as the input data \cite{pseudolidar,am3d,monowithpl,decoupled3d,patchnet,demystifying,pct,neighborvote,da3ddet,foresee} or augment the input images \cite{multifusion,d4lcn,ddmp3d,dfrnet,roi10d}. Note this scheme may lead to data leakage in KITTI 3D.

\noindent
{\bf Data leakage.}
As stated above, some methods choose to pre-train a sub-network on the KITTI Depth or KITTI Stereo. Unfortunately, there is an overlap between the training set of KITTI Depth/Stereo and the validation set of KITTI 3D, which may cause the data leakage problem. Wang {\it et al.} \cite{pseudolidar} aware of this problem, propose to re-train their disparity estimation model from the images provided by the training split of the KITTI 3D, but this issue in monocular depth estimators still remains. Even worse is the fact that almost all the monocular 3D detectors \cite{pseudolidar,am3d,d4lcn,multifusion,decoupled3d,monowithpl,ddmp3d,da3ddet,patchnet,neighborvote,dfrnet,pct} with pre-computed depth maps inherit this problem, which leads to the unfair and unreliable comparison on the KITTI 3D validation set. Recently, Simonelli {\it et al.} \cite{demystifying} revisit this issue, and provide a new training/validation split which avoids the overlap by comparing the GPS data corresponding to each image. Unfortunately, their experimental results show this issue cannot been completely fixed, and future works are recommended to consider this problem when building their models.

\noindent
{\bf Instance segmentor.}
Instance mask plays an important role in many image-based 3D detectors such as \cite{monowithpl,dispnet,zoomnet,oc_stereo,monocinis,car_parsing}, while the ground truth is not available from the 3D detection datasets. Except for generating labels using LiDAR signals (Section \ref{sec:auxdata_lidar}), another choice is to pre-train an instance segmentation from external datasets, {\it e.g.} CityScapes \cite{cityscapes} which contains 5,000 images with pixel-wise annotation in the autonomous driving scenario. 
Then we can fix the weights of the segmentation network and use it to predict masks in the detection system \cite{monowithpl, monocinis}.

\subsubsection{Backbone Pre-Training}

Except for leveraging fixed pre-trained models, some works also pre-train their backbones on external datasets (excluding ImageNet pre-training) for better weight initialization.

\noindent
{\bf Pre-training for monocular models.}
Park {\it et al.} \cite{dd3d} present that depth estimation can serve as a proxy task for backbone pre-training and provide rich geometric priors to the networks. Particularly, they pre-train their model on the large-scale DDAD15M dataset \cite{ddad} which contains about 15M images of urban driving scenes for depth estimation. Compared with training from scratch and fine-tuning from ImageNet pre-trained model, this pre-training strategy significantly boosts the accuracy of monocular 3D detectors and has been confirmed by several works \cite{bevdet, detr3d}.

\noindent
{\bf Pre-training for stereo models.}
The stereo-matching model is generally involved in 3D detectors as a sub-model. To improve the accuracy of stereo matching, Wang {\it et al.} \cite{pseudolidar} pre-train their stereo network from Scene Flow \cite{sceneflow}, which provides more than 30K synthetic stereo pairs with dense annotation. This choice has been adopted by some following works \cite{zoomnet,oc_stereo,patchnet,pseudolidar++,pseudolidare2e}.



\begin{table*}
\centering
\caption{The requirements of auxiliary data and results of the images-based 3D detectors published on top-tier conferences and journals. We report  $\rm{AP}|_{R_{40}}$ for 3D/BEV detection on the KITTI {\it test} set of the Car category.
We group the methods according to the auxiliary data they used, and the methods in each group are ranked by the 3D $\rm{AP}|_{R_{40}}$ under the moderate setting. We also show whether the methods use the pre-training weights (excluding ImageNet pre-training).}
\resizebox{0.99\linewidth}{!}{
\begin{tabular}{|ll|c|ccc|ccc|}
\hline
\multirow{2}{*}{Methods} & 
\multirow{2}{*}{Venue} & 
\multirow{2}{*}{Pre-Train} & 
\multicolumn{3}{c}{Auxiliary Data} &
\multicolumn{3}{|c|}{Performance} \\ 
\cline{4-6}  \cline{7-9}
 ~ & ~ & ~ & LiDAR & Temporal & Stereo & Easy & Mod. & Hard \\
\hline
3DVP\cite{3dvp} & \emph{CVPR'15} & - & - & - & - &
- / - & - / - & - / -  \\
Mono3D\cite{mono3d} & \emph{CVPR'16} & \checkmark & - & - & -  & 
- / - & - / - & - / -  \\
Deep3DBox\cite{deep3dbox} & \emph{CVPR'17} & - & - & - & - & 
- / - & - / - & - / -\\
Deep MANTA\cite{deepmanta} & \emph{CVPR'17} & - & - & - & - &  - / - & - / - & - / - \\
Multi-Fusion\cite{multifusion} & \emph{CVPR'18} & \checkmark & - & - & - &
- / -  & - / - & - / - \\
Mono3D++\cite{mono3d++} & \emph{AAAI'19} & \checkmark & - & - & - & 
- / - & - / - & - / -  \\
Chang {\it et al.}\cite{deepoptics} & \emph{ICCV'19} & \checkmark & - & - & - & 
- / - & - / - & - / -  \\
ForeSeE\cite{foresee} & \emph{AAAI'20} & \checkmark & - & - & - &
- / - &  - / -  & - / - \\
MoNet3D\cite{monet3d} & \emph{ICML'20} & - & - & - & - & 
- / - & - / - & - / -\\
EgoNet\cite{egonet} & \emph{CVPR'21} & - & - & - & - & 
- / - & - / - & - / -\\
FQNet\cite{fqnet} & \emph{CVPR'19} & - & - & - & - & 
 2.77 / 5.40 & 1.51 / 3.23 & 1.01 / 2.46 \\
ROI-10D\cite{roi10d} & \emph{CVPR'19} & \checkmark & - & - & - &
4.32 / 9.78 & 2.02 / 4.91 & 1.46 / 3.74\\
GS3D\cite{gs3d} & \emph{CVPR'19} & - & - & - & - & 
4.47 / 8.41 & 2.90 / 6.08 & 2.47 / 4.94\\
MonoFENet\cite{monofenet} & \emph{T-IP'19}  & \checkmark & - & - & - &
8.35 / 17.03 &  5.14 / 11.03  & 4.10 / 9.05 \\
MonoGRNet\cite{monogrnet} & \emph{AAAI'19} & - & - & - & - & 
9.61 / 18.19 & 5.74 / 11.17 & 4.25 / 8.73 \\
Decoupled-3D\cite{decoupled3d} & \emph{AAAI'20}  & \checkmark & - & - & - &
11.08 / 23.16 & 7.02 / 14.82  &  5.63 / 11.25 \\
MonoDIS\cite{monodis} & \emph{ICCV'19} & - & - & - & - & 
 10.37 / 17.23  & 7.94 / 13.19 &  6.40 / 11.12  \\
UR3D\cite{ur3d} & \emph{ECCV'20} & - & - & - & - & 
15.58 / 21.85 & 8.61 / 12.51 & 6.00 / 9.20 \\
Neighbor-Vote\cite{neighborvote} & \emph{ACM MM'21} & \checkmark & - & - & - &
15.57 / 27.39 & 9.90 / 18.65 & 8.89 / 16.54 \\
M3D-RPN\cite{m3drpn} & \emph{ICCV'19} & - & - & - & - & 
14.76 / 21.02 & 9.71 / 13.67 & 7.42 / 10.23\\
MonoPair\cite{monopair} & \emph{CVPR'20} & - & - & - & - & 
13.04 / 19.28 & 9.99 / 14.83 & 8.65 / 12.89  \\
RTM3D\cite{rtm3d} & \emph{ECCV'20} & - & - & - & - & 14.41 / 19.17 & 10.34 / 14.20 & 8.77 / 11.99  \\
AM3D\cite{am3d} & \emph{ICCV'19} & \checkmark & - &  - & - &
16.50 / 25.03 & 10.74 / 17.32  & 9.52 / 14.91  \\
MoVi-3D\cite{movi3d} & \emph{ECCV'20} & - & - & - & - & 15.19 / 22.76 & 10.90 / 17.03 & 9.26 / 14.85  \\
RAR-Net\cite{rarnet} & \emph{ECCV'20} & - & - & - & - & 
16.37 / 22.45  & 11.01 / 15.02 & 9.52 / 12.93 \\
PatchNet \cite{patchnet} & \emph{ECCV'20} & \checkmark & - & - & - &
15.68 / 22.97 & 11.12 / 16.86 & 10.17 / 14.97\\
M3DSSD\cite{m3dssd} & \emph{CVPR'21} & - & - & - & - & 
 17.51 / 24.15 & 11.46 / 15.93 &  8.98 / 12.11  \\
 D4LCN\cite{d4lcn} & \emph{CVPR'20} & \checkmark & - & - & - &
16.65 / 22.51 & 11.72 / 16.02 &  9.51 / 12.55 \\
MonoDLE\cite{monodle} & \emph{CVPR'21}  & - & - & - & - & 
17.23 / 24.79 & 12.26 / 18.89 & 10.29 / 16.00 \\
GrooMeD-NMS\cite{groomednms} & \emph{CVPR'21}  & - & - & - & - & 
18.10 / 26.19 & 12.32 / 18.27 & 9.65 / 14.05\\
Demystifying\cite{demystifying} & \emph{ICCV'21} & \checkmark & - & - & - & 
22.40 / - &  12.53 / - & 10.64 / - \\
Mono R-CNN\cite{monorcnn} & \emph{ICCV'21}  & - & - & - & - & 
18.36 / 25.48 &  12.65 / 18.11 & 10.03 / 14.10 \\
DDMP-3D\cite{ddmp3d} & \emph{CVPR'21} & \checkmark & - & - & - &
 19.71 / 28.08 & 12.78 / 17.89 &  9.80 / 13.44\\
MonoDIS (multi)\cite{monodis_journal} & \emph{T-PAMI'20} & - & - & - & - & 
16.54 / 24.45 &  12.97 / 19.25 & 11.04 / 16.87 \\
PCT \cite{pct} & \emph{NeurIPS'21} & \checkmark & - & - & - &
21.00 / 29.65 & 13.37 / 19.03 & 11.31 / 15.92  \\
DFRNet\cite{dfrnet} & \emph{ICCV'21} & \checkmark & - & - & - & 
 19.40 / 28.17 & 13.63 / 19.17 & 10.35 / 14.84 \\
MonoEF\cite{monoef} & \emph{CVPR'21} & \checkmark & - & - & - &
21.29 / 29.03 &  13.87 / 19.70 &  11.71 / 17.26 \\
MonoFlex\cite{monoflex} & \emph{CVPR'21}  & - & - & - & - & 
19.94 / 28.23 & 13.89 / 19.75 & 12.07 / 16.89 \\
AutoShape\cite{autoshape} & \emph{ICCV'21} & - & - & - & - & 
 22.47 / 30.66 &  14.17 / 20.08 &  11.36 / 15.95 \\
GUPNet\cite{gupnet} & \emph{ICCV'21} & - & - & - & - & 
20.11 / 30.29 & 14.20 / 21.19 &  11.77 / 18.20 \\
DD3D\cite{dd3d} & \emph{ICCV'21} & \checkmark & - & - & - &
 23.22 / 32.35 & 16.34 / 23.41 & 14.20 / 20.42 \\
MonoCon\cite{monocon} & \emph{AAAI'22} & - & - & - & - &
22.50 / 31.12 & 16.46 / 22.10 & 13.95 / 19.00  \\
\hline
MonoPSR\cite{monopsr} & \emph{CVPR'19} & - & \checkmark & - & - &
 10.76 / 18.33 & 7.25 / 12.58 & 5.85 / 9.91\\
DA-3Ddet\cite{da3ddet} & \emph{ECCV'20} & \checkmark & \checkmark & - & - &
16.77 / - & 11.50 / - & 8.93 / - \\
MonoRUn\cite{monorun} & \emph{CVPR'21}  & - & \checkmark & - & - &
19.65 / 27.94 & 12.30 / 17.34 &  10.58 / 15.24 \\
CaDDN\cite{caddn} & \emph{CVPR'21}  & - & \checkmark & - & - & 
19.17 / 27.94 & 13.41 / 18.91 &   11.46 / 17.19 \\
MonoDistill\cite{monodistill} & \emph{ICLR'22} & - & \checkmark & - & - & 
22.97 / 31.87 & 16.03 / 22.59 &  13.60 / 19.72 \\
\hline
Kinematic3D\cite{kinematic3d} & \emph{ECCV'20} & - & - & \checkmark & - &
 19.07 / 26.69 & 12.72 / 17.52 & 9.17 / 13.10 \\
\hline
3DOP\cite{3dop} & \emph{NeurIPS'15} & - & - & - & \checkmark &
- / - & - / - & - / - \\
3DOP\cite{3dop_journal} & \emph{T-PAMI'17} & - & - & - & \checkmark &
- / - & - / - & - / - \\
IDA-3D\cite{ida3d} & \emph{CVPR'20} & - & - & - & \checkmark &
- / -  & - / - & - / - \\
TLNet\cite{tlnet} & \emph{CVPR'19} & - & - & - & \checkmark &
 7.64 / 13.71 & 4.37 / 7.69 & 3.74 / 6.73 \\
Stereo R-CNN\cite{stereorcnn} & \emph{CVPR'19} & - & - & - & \checkmark &
 47.58 / 61.92 & 30.23 / 41.31 & 23.72 / 33.42 \\
RTS3D\cite{rts3d} & \emph{AAAI'21} & - & - & - & \checkmark & 
 58.51 / 72.17 &  37.38 / 51.79 & 31.12 / 43.19 \\
Disp R-CNN\cite{disprcnn} & \emph{CVPR'20}  & - & - & - & \checkmark &
 58.53 / 73.82 & 37.91 / 52.34 & 31.93 / 43.64 \\
Disp R-CNN\cite{disprcnn_j} & \emph{T-PAMI'21} & - & - & - & \checkmark &
 67.02 / 79.61 & 43.27 / 57.98 & 36.43 / 47.09 \\
\hline
Pseudo-LiDAR\cite{pseudolidar} & \emph{CVPR'19}  & \checkmark  & \checkmark & - & \checkmark & 
54.53 / 67.30 & 34.05 / 45.00 & 28.25 / 38.40 \\
ZoomNet\cite{zoomnet} & \emph{AAAI'20} & \checkmark & \checkmark & - & \checkmark & 
55.98 / 72.94 & 38.64 / 54.91 & 30.97 / 44.14\\
Pseudo-LiDAR++\cite{pseudolidar++} & \emph{ICLR'20} & \checkmark & \checkmark & - & \checkmark &
61.11 / 78.31 &  42.43 / 58.01 &  36.99 / 51.25 \\
E2E Pseudo-LiDAR\cite{pseudolidare2e} & \emph{CVPR'20} & \checkmark & \checkmark & - & \checkmark &
64.8 / 79.6 &  43.9 / 58.8  & 38.1 / 52.1 \\
DSGN\cite{dsgn} & \emph{CVPR'20} & - & \checkmark & - & \checkmark &
 73.50 / 82.90 & 52.18 / 65.05 &  45.14 / 56.60 \\
CDN\cite{wasserstein} & \emph{NeurIPS'20} & - & \checkmark & - & \checkmark &74.52 / 83.32 & 54.22 / 66.24 & 46.36 / 57.65 \\
LIGA Stereo\cite{ligastereo} & \emph{ICCV'21} & - & \checkmark & - & \checkmark &81.39 / 88.15 & 64.66 / 76.78 & 57.22 / 67.40  \\
\hline
\end{tabular}}
\label{table:auxdata}
\end{table*}

\subsection{Input Data as Taxonomy}
In the above sections, we summarize the main applications of the auxiliary data. Note the benefits of these data to the 3D detectors are also different, and it is unfair to compare the methods without considering the data they use. Consequently, we argue that the underlying input data of algorithms can also serve as the taxonomy, and the comparisons should be conducted under the same setting.
Here we group these methods according to the auxiliary data ({\it i.e.} LiDAR signals, temporal sequences, and stereo pairs) they used, and show the results achieved by existing methods on the most commonly used KITTI 3D dataset and the auxiliary data they used in Table \ref{table:auxdata}.
From these results, we can find the following observations: (i) Most methods (39 of 63 in Table \ref{table:auxdata}) adopt at least one kind of auxiliary data (or pre-training weights) in their models, which shows the wide application of auxiliary data and suggests estimating 3D bounding boxes from a single image is full of challenge. (ii) The stereo images provide the most valuable information for image-based 3D detection, and the methods with this kind of data at the inference stage perform significantly better than others. (iii) Although the temporal visual features in video are vital to the visual perception system, there is only one method \cite{kinematic3d} leveraging the temporal sequences, and more attention is encouraged to using this kind of data in the future work. (iv) The performance of existing methods has been rapidly and constantly improved. Take the methods without any auxiliary data as an example, the SOTA performance on the KITTI benchmark (moderate setting) has increased to 16.46 (MonoCon\cite{monocon}, published on AAAI'22) from 1.51 (FQNet \cite{monogrnet}, published on CVPR'19)


We hope the analyses and numbers in this section can provide a clear presentation for existing methods and fair baselines for future works. Due to the space limitation, we only present the methods published on the most relevant conferences (CVPR, ICCV,  ECCV, AAAI, ICLR, ICML, and NeurIPS) and their journal versions in Table \ref{table:auxdata}. More information, including the methods published on other venues, can be found in this website: \url{https://github.com/xinzhuma/3dodi-survey}.

%% file: components/future_direction.tex
\section{Future Directions}
\label{sec:future}
Although the detection accuracy of the detectors has been rapidly and constantly improving, image-based 3D object detection is still a relatively new field and there are many limitations and directions which need to be further analyzed and explored. \revision{Generally, the main challenges come from two aspects: ({\it i}): perceiving the 3D world from 2D images is an ill-posed problem; and ({\it ii}): labeling 3D data is extremely expensive and the scale of existing datasets is still limited. For the first issue, exploring better \emph{depth estimation} methods is a promising research direction, and leveraging \emph{multi-modality} and \emph{temporal sequences} can also alleviate this problem. For the second one, \emph{methods beyond fully supervised learning} should be encouraged. Besides, \emph{transfer learning from (large) pre-trained models} is also a reasonable solution. Lastly, considering the application scenarios of this task, the \emph{generalization ability} of the models is an issue that cannot be ignored. We provide more discussions for these topics in the following parts, hoping to provide relevant cues for impactful future work.}

\subsection{Depth Estimation}
The performance of image-based 3D object detection methods heavily relies on the capability of estimating the precise distance of the objects. A relevant future direction is therefore to analyze and improve the depth estimation capabilities of 3D object detectors. Many recent works, such as \cite{caddn,monorcnn,gupnet,wasserstein,foresee,pgd}, try to address this, proposing alternative definitions for the regression targets and loss formulations and demonstrating that there is still a lot of room for improvement.

Another interesting future direction comes by observing that, quite surprisingly, the depth estimation and 3D object detection communities have been almost completely independent from one another. A first attempt to join these communities has been made with the introduction of Pseudo-LiDAR methods \cite{pseudolidar,am3d,monowithpl}, where 3D object detectors have been paired with pre-trained depth estimators and demonstrated to achieve better overall performance. While this is a promising initial step, the depth and detection methods were still completely independent. To overcome this, \cite{pseudolidare2e, isplneeded} proposed to join the 3D detection and depth estimation into a single multi-task network. They demonstrated that, when these two tasks are trained together and have the possibility to benefit from one another, the 3D detection performance increases even more. We believe these results show and validate the potential of the union of depth and detection, highlighting that this will constitute a relevant future direction.

\subsection{Multi-Modality}
For the 3D object detection task, both image data and LiDAR data have their advantages (see Appendix \ref{sec:image_vs_lidar} for the discussion of these two kinds of data), and some methods, such as \cite{mv3d,avod,frustumpointnet,contfuse}, have recently started to integrate these two types of data into a single model. However, the research in this field is still on its infancy. Additionally, other modalities of data could be considered to further improve the accuracy and robustness of algorithms. For example, compared with LiDAR, RADAR equipment has a longer sensing distance, which may be used to boost the accuracy of far-away objects. Besides, RADAR is more stable in some extreme weather conditions, such as rainy day and foggy day. However, although the synchronized RADAR data are already provided in some datasets~\cite{nuscenes,multisensorfusion}, there are only a few methods \cite{multisensorfusion,lidar_radar,centerfusion} which investigate how to use them. Another example is the data from thermal cameras \cite{kaist}, which provides new opportunities to advance detection accuracy by tackling adverse illumination conditions. In summary, the ideal detection algorithms should integrate a variety of data, to cover heterogeneous and extreme conditions.

\subsection{Temporal Sequences}
In the real world, human drivers rely on \emph{continuous} visual perception to obtain information about the surrounding environment. However, most of the works in this field solve the 3D detection problem from the perspective of a single frame, which is obviously sub-optimal, and only one recently work~\cite{kinematic3d} has started to consider temporal cues and constraints. On the other hand, lots of works had proved the effectiveness of using video data in many tasks, including 2D detection~\cite{video_detection1,video_detection2}, depth estimation~\cite{video_depth1,video_depth2}, and LiDAR-based 3D detection \cite{faf,hindsight}. The successes in these related fields demonstrate the potential of leveraging video data in the 3D detection task, and new breakthroughs can be achieved by introducing the temporal data and building new constraints in the spatio-temporal space.

A particularly interesting future direction regarding the use of sequences is that they can be used to relax the requirement of full-supervision. If combined with the already available input RGB images in fact, they are demonstrated to enable self-supervised depth estimation \cite{packnet}. In light of this, it is reasonable to think that if the same supervision would be used to also recover the shape and appearance of objects, the same approach could be used to perform 3D object detection as suggested by \cite{Ost_2021_CVPR, monodr}. 

A last relevant direction is represented by velocity estimation. Some datasets, {\it e.g.} nuScenes \cite{nuscenes}, are in fact required to estimate not only the 3D boxes of objects but also their velocities {\it w.r.t.} the global coordinate system. This introduces another extremely challenging task that requires to be solved through the use of multiple images.

\subsection{Beyond Fully Supervised Learning}\label{sec:beyondfullysup}
The creation of 3D detection datasets is an extremely expensive and time-consuming operation. It generally involves the synergy between different technologies ({\it e.g.} LiDAR, GPS, cameras) as well as a substantial amount of workforce. The annotation process is highly demanding and, even in the presence of many quality checks, it is inevitably affected by errors, especially for long-range objects. In light of this, it is concerning to see that the almost totality of the 3D object detection methods is fully supervised, {\it i.e.} requires the 3D bounding box annotations to be trained. Contrarily to other related communities where the full supervision requirement has been relaxed {\it e.g.} depth estimation \cite{monodepth,monodepth2} or LiDAR-based 3D detection \cite{wypr,sess}, very little effort has been devoted to exploring semi- or self-supervised approaches \cite{monodr,geo_semi,weakm3d}. In this regard, it is worth to highlight the method in \cite{monodr}, which introduces a differentiable rendering module that enables to exploit input RGB image as the only source of supervision. Also in light of the recent advancements in the field of differentiable rendering on generic scenes ({\it e.g.} NeRF \cite{mildenhall2020nerf}) and real objects ({\it e.g.} \cite{Niemeyer2020GIRAFFE}, \cite{Ost_2021_CVPR}), we believe this particular direction to be extremely valuable and able to potentially relax the requirements of 3D box annotations.
Besides, to address possible errors in data annotations, {\it e.g.} missing annotation for long-range objects, the geometry consistency between temporal sequences or multi-frame is also encouraged for both full-, semi-, or self-supervised learning.

\subsection{Pre-Training}
As discussed in Section \ref{sec:pretraining}, some works pre-train certain components of the model. However, the vast majority of the methods still train their models from scratch or use ImageNet pre-trained weights. We expect approaches that adopt pre-training to be further investigated and become more popular, especially taking into account their potential in challenging scenarios, {\it e.g.} unsupervised settings. Particularly, related techniques, such as BERT \cite{bert} or MoCo \cite{moco}, achieve great success in NLP and 2D vision, while have not been introduced in to image-based 3D detection field. Such technologies may be able to leverage the massive un-labeled data in the autonomous driving scenario and further boost the accuracy of 3D detectors.

\subsection{Generalization}
Generalization plays an important role in the security of self-driving cars. In this regard, it is unfortunately quite well known that image-based 3D object detection methods experience a quite significant drop in performance when tested on unseen datasets, objects, or challenging weather conditions. An example can be found in Table~\ref{table:robustness}, where we show the performance of an image-based baseline (along with a LiDAR baseline) on subsets of the popular nuScenes dataset which contain images captured with rain or at night. On the many factors that cause this performance drop, there is certainly the issue that almost the totality of image-based 3D detectors is camera dependent {\it i.e.} they expect the camera intrinsic parameters to be unchanged between the training and testing phase. Initial attempts to overcome this limitation have been developed in \cite{monocinis} but we believe that this direction should be further explored. Another crucial factor comes from the fact that many image-based 3D object detection methods rely on dataset specific object priors {\it i.e.} average object 3D extents in order to make their predictions. If tested on different datasets where the objects, {\it e.g.} cars, are significantly deviating from these average extents then the 3D detector is likely to fail. Since the effort towards solving this issue have very limited~\cite{trainingermany, st3d, spg,mlcnet} and 
uniquely focused on LiDAR-based approaches, we believe that this also constitutes a relevant future direction.

%% file: components/conclusion.tex
\section{Conclusions}
\label{sec:conclusion}

This paper provides a comprehensive survey of the recent developments in image-based 3D detection for autonomous driving.
We have seen that, from 2015 to 2021, lots of papers on this topic have been published. To summarize these methods systematically, we first give a taxonomy of existing methods according to their high-level structure. Then, a detailed comparison of these algorithms is given, discussing each necessary component for 3D detection, such as feature extraction, loss formulation, post-processing, etc. We also discuss the applications of auxiliary data in this field, supporting the need for a systematic summary as in this survey and better protocol for fair comparisons in future work. Finally, we describe some of the open challenges and potential directions in this field that could spur new research in the coming years.

%% file: components/appendix.tex
\newpage
\setcounter{section}{0}
\renewcommand\thesection{\Alph{section}} 
\section{Appendix}

This document introduces the evaluation metrics of 3D object detection in Section \ref{supp:metric}, summarize the NMS algorithms in Section \ref{supp:nms}, and discuss the relevant problem about image-based 3D detection in Section \ref{supp:discussion}.

\subsection{Evaluation Metrics}
\label{supp:metric}
As for 2D object detection, the Average Precision (AP) ~\cite{ap,voc} constitutes the main evaluation metric used in 3D object detection. Starting from its vanilla definition, each dataset has applied specific modifications which gave rise to dataset specific evaluation metrics. Here, we first review the original AP metric, and then introduce its variants adopted in the most commonly used benchmarks, including the KITTI 3D, nuScenes, and Waymo Open.

\subsubsection{Review of the AP Metric}
To compute AP, the predictions are first assigned to their corresponding ground truths according to a specific measure. The most commonly used one, {\it i.e.} the Intersection over Union (IoU) between the ground truth $A$ and the estimated 3D bounding box $B$, is defined as:
\begin{equation}
{\rm IoU}(A, B) = \frac{|A\cap B|}{|A\cup B|}.
\label{equ:iou}
\end{equation}
The IoU measure is used to judge a matched prediction as a True Positive (TP) or a False Positive (FP) by comparing it with a certain threshold.  Then, the recall $r$ and precision $p$ can be computed from the ranked (by confidence) detection results according to:
\begin{equation}
r = \frac{\rm{TP}}{\rm{TP} + \rm{FN}}, \quad p = \frac{\rm{TP}}{\rm{TP} + \rm{FP}},
\label{equ:recall&precision}
\end{equation}
where the FN denotes the False Negative. The precision can be regarded as a function of recall, {\it i.e.} $p(r)$. Furthermore, to reduce the impact of "wiggles" in the precision-recall curve~\cite{voc,info_retrieval}, the interpolated precision values are used to  compute the AP using:
\begin{equation}
{\rm AP} = \frac{1}{|\mathbb{R}|} \sum_{r\in \mathbb{R}} p_{interp}(r),
\label{equ:ap}
\end{equation}
where $\mathbb{R}$ is the predefined set of recall positions and $p_{interp}(r)$ is the interpolation function defined as :
\begin{equation}
p_{interp}(r) = \max_{r':r'\geq r} p(r'),
\label{equ:interp}
\end{equation}
which means that instead of averaging over the actually observed precision values at recall $r$, the maximum precision at recall value greater than or equal to $r$ is taken.

\subsubsection{Dataset Specific Metrics}
{\bf KITTI 3D Benchmark.} 
KITTI 3D adopts the AP as the main metric and introduces some modifications. The first one is that the computation of the IoU is done in 3D space. Besides, KITTI 3D adopted the suggestion of Simonelli {\it et al.} \cite{monodis} and replaced $\mathbb{R}_{11}=\{0, 1/10, 2/10, 3/10, ..., 1\}$ in Equation~\ref{equ:ap} with $\mathbb{R}_{40}=\{1/40, 2/40, 3/40, ..., 1\}$, which is a more dense sampling with the removal of recall position at 0.

Furthermore, due to the height of the objects is not so important as other items in the autonomous driving scenarios, Bird's Eye View (BEV) detection, also known as 3D localization task in some works \cite{mv3d,frustumpointnet,am3d}, can be seen as an alternative to 3D detection. The calculation process of the metric, BEV AP, for this task is the same as the 3D AP, but the IoU is calculated on the ground plane, instead of the 3D space. This task is also included in some other benchmarks, such as Waymo Open \cite{waymo}.

Besides, KITTI 3D also proposed a new metric, Average Orientation Similarity (AOS), to evaluate the accuracy of orientation estimation. AOS is formulated as:
\begin{equation}
{\rm AOS} = \frac{1}{|\mathbb{R}|} \sum_{r\in \mathbb{R}} \max_{r':r'\geq r} s(r').
\label{equ:aos}
\end{equation}
The orientation similarity $s(r)\in [0, 1]$ in Equation~\ref{equ:aos} for recall $r$ is a normalized variant of the cosine similarity deﬁned as:
\begin{equation}
s(r) = \frac{1}{|D(r)|}\sum_{i\in D(r)} \frac{1+\cos\Delta_{\theta}^{(i)}}{2}\delta_{i},
\label{equ:os}
\end{equation}
where $D(r)$ denotes the set of all object detection results at recall rate $r$ and $\Delta_{\theta}^{(i)}$ is the difference in angle between the estimated and ground-truth orientations of detection $i$. To penalize multiple detections for a single object, KITTI 3D enforces $\delta_{i}=1$ if detection $i$ has been assigned to a ground-truth bounding box and $\delta_{i}=0$ if it has not been assigned.
Note that all the AP metrics are computed independently for each difficulty level and category.

\noindent
{\bf Waymo Open Benchmark.}
Waymo Open also adopted the AP metric with a minor modification: replacing $\mathbb{R}_{11}$ in Equation~\ref{equ:ap} with $\mathbb{R}_{21}=\{0, 1/20, 2/20, 3/20, ..., 1\}$. Moreover, considering that accurate heading prediction is critical for autonomous driving and the AP metric does not have a notion of heading, Waymo Open further proposes Average Precision weighted by Heading (APH) as its primary metric. Specifically, APH incorporates heading information into the precision calculation. Each true positive is weighted by the heading accuracy deﬁned as $\min(|\theta-\theta^{*}|, 2\pi-|\theta-\theta^{*}|)/\pi$, where $\theta$ and $\theta^{*}$ are the predicted heading angle and the ground-truth heading angle in radians within $[-\pi,\pi]$. Note that APH jointly assesses the performance of both {\it 3D detection} and {\it orientation estimation}, while AOS is only designed for {\it orientation estimation}.

\noindent
{\bf nuScenes Benchmark.}
nuScenes proposed a new AP-based metric. In particular, it uses the 2D center distance on the ground plane to match the predictions and ground truths with a certain distance threshold $d$ ({\it e.g.} 2m), instead of the IoU introduced in Equation~\ref{equ:iou}. Besides, nuScenes calculate AP as the normalized area under the precision-recall curve for recall and precision over 10\%. Finally, it calculates the mean Average Precision (mAP) over matching thresholds of $\mathbb{D}=\{0.5,1,2,4\}$ meters and the set of classes $\mathbb{C}$:
\begin{equation}
{\rm mAP} = \frac{1}{|\mathbb{C}||\mathbb{D}|}\sum_{c \in \mathbb{C}}\sum_{d\in \mathbb{D}}{\rm AP}_{c, d}.
\label{equ:map_nuscenes}
\end{equation}
However, this metric only considers the localization of the objects, ignoring the effects of other aspects such as dimension and orientation. To compensate for it, nuScenes also proposed a set of True Positive metrics (TP metrics) designed to measure each predicted error separately using all true positives (determined under the center distance $d = 2m$ during matching). All the five TP metrics are designed to be positive scalars, which are defined as follows \cite{nuscenes}:
\begin{itemize}
\item \emph{Average Translation Error (ATE)} is the Euclidean distance for object center on the 2D ground plane (units in meters).
\item \emph{Average Scale Error (ASE)} is the 3D IoU error $(1-{\rm IoU})$ after aligning orientation and translation.
\item \emph{Average Orientation Error (AOE)} is the smallest yaw angle difference between the predictions and ground truths (in radians). 
\item \emph{Average Velocity Error (AVE)} is the absolute velocity error as the L2 norm of the velocity differences in 2D (in m/s).
\item \emph{Average Attribute Error (AAE)} is deﬁned as 1 minus attribute classiﬁcation accuracy $(1-acc)$. 
\end{itemize}
Furthermore, for each TP metric, nuScenes also computes the mean TP metric (mTP) over all object categories:
\begin{equation}
{\rm mTP}_k = \frac{1}{|\mathbb{C}|} \sum_{c \in \mathbb{C}} \mathrm{TP}_{k, c},
\label{equ:mean_tp_metrics}
\end{equation}
where $\mathrm{TP}_{k, c}$ denotes the $k^{th}$ TP metric ({\it e.g.} k = 1 means the ATE) for category $c$. Finally, to integrate all the mentioned metrics to a scalar score, nuScenes further proposes the nuScenes Detection Score (NDS) that combines the mAP defined in Equation~\ref{equ:map_nuscenes} and the mTP${_k}$ defined in Equation~\ref{equ:mean_tp_metrics}:
\begin{equation}
{\rm NDS} = \frac{1}{10} [5\cdot {\rm mAP}+\sum_{k=1}^5(1-\min(1, {\rm mTP}_{k}))].
\label{equ:mean_tp_metrics2}
\end{equation}

\subsection{NMS}
\label{supp:nms}
\noindent
{\bf Traditional NMS.}
Generally, the original detection results have multiple redundant bounding boxes covering a single object, and NMS is designed so that a single object is only covered by an estimated bounding box. The pseudo-code for the traditional NMS is shown in Algorithm \ref{alg:nms}.
\begin{algorithm}
\caption{Traditional NMS}
\label{alg:nms}
\KwData{$\mathcal{B}=\left\{b_{1}, \ldots, b_{n}\right\}$, $\mathcal{S}=\left\{s_{1}, \ldots, s_{n}\right\}$, $\Omega$ \\ $\mathcal{B}$ is the list of initial bounding boxes,  $\mathcal{S}$ contains corresponding scores, $\Omega$ is the NMS threshold} 

\KwResult{$\mathcal{D}$, the set of final results with scores}

$\mathcal{D} \leftarrow \varnothing$\;
\While{$\mathcal{S} \neq \varnothing$}{
$m \leftarrow\arg\max \mathcal{S}$\;
$\mathcal{B} \leftarrow \mathcal{B} - \left\{b_{m}\right\}$\;
$\mathcal{S} \leftarrow \mathcal{S} - \left\{s_{m}\right\}$\;
$\mathcal{D} \leftarrow \mathcal{D} \cup \left\{(b_{m}, s_{m})\right\} $\;
\For{$b_{j}\in \mathcal{B}$}{
\If{$\mathrm{IoU}(b_{m}, b_{j}) > \Omega$}{
$\mathcal{B} \leftarrow \mathcal{B}- \left\{b_{m}\right\}$\;
$\mathcal{S} \leftarrow \mathcal{S}- \left\{s_{m}\right\}$\;}
}}
\Return $\mathcal{D}$;
\end{algorithm}
In particular, the bounding box $b_{m}$ with the maximum score is selected and all other boxes having high overlap with $b_{m}$ are removed from the detection results. This process is recursively applied on the remaining boxes to get the final results.

\noindent
{\bf The variants of NMS.}
To avoid the removal of valid objects, Bodla {\it et al.} \cite{softnms} just reduce the scores of high overlapped objects, instead of discarding them (Soft NMS). Jiang {\it et al.} \cite{iounet} observe the mismatch between the classification score and the quality of box, and propose to regress a localization score, {\it i.e.} IoU score, to play the role of $\mathcal{S}$ in Algorithm \ref{alg:nms} (IoU Guided NMS).  Since the major issue of monocular based 3D detectors is the localization error \cite{monodle,oftnet}, where the depth estimation is the core problem to recover object location, Shi {\it et al.} \cite{ur3d} use Equation~\ref{equ:laplacian} to capture the uncertainty of estimated depth, and use the depth uncertainty $\sigma_{depth}$ to normalize the  score $s$ to $\frac{s}{\sigma_{depth}}$ when applying the NMS (Depth Guided NMS). It is reported in \cite{deepid,weightednms} that the boxes with non-maximum scores may also have high-quality localization and propose to update $b_{m}$ by weighted averaging of the boxes $b_{i}$ with high overlap (Weighted NMS). In particular, they first compute weight by: $w_{i} = s_{i} \times \mathrm{IoU}(b_{m}, b_{i})$ and update the bounding box by: $b_{i} = \sum_{i}\frac{w_{i}}{sum(w)} \cdot b_{i}$, where $sum(w) = \sum_{i} w_{i}$. Similarly, He {\it et al.} \cite{softernms} also adopt weighted averaging mechanism with an update of the averaging rule. Particularly, they model the uncertainty of each item of bounding box under Gaussian distribution (Equation \ref{equ:gaussian}) and then set the averaging rule only related to the IoU and uncertainty (Softer NMS). Liu {\it et al.} \cite{adaptivenms} propose to use a dynamic NMS threshold $\Omega$ for objects with different densities (Adaptive NMS).

Note that some algorithms mentioned above \cite{softnms,softernms,weightednms,adaptivenms,iounet,deepid} are initially proposed for 2D detection, but they can be easily applied to 3D detection. Besides, \cite{weightednms,softernms} can also be regarded as post-optimization methods because they update the predicted results during the NMS process, except for eliminating duplicated detections.

\noindent
{\bf Others.} 
Kumar {\it et al.} \cite{groomednms} propose a differentiable NMS for monocular 3D detection. With this design, the loss function can directly operate on the results after NMS. 
Besides, for the multi-camera-based panoramic datasets, {\it e.g.} nuScenes and Waymo Open, the global NMS is needed to eliminate the duplicate detection results from the overlapping images.

\input{components/dicussion}

%% file: components/dicussion.tex
\subsection{More Discussions}
\label{supp:discussion}
In this section, we provide additional discussions about the image-based 3D detection, including the metrics it uses, the trade-off between accuracy and speed, \revision{the comparisons with LiDAR-based methods, and the discussion for image, pseudo-LiDAR, and LiDAR representations.}

\begin{table*}[!t]
\caption{Detailed performances of  PointPillars/MonoDIS on the nuScenes \emph{test} set. 
We use \textcolor{red}{red}/\textcolor{blue}{blue} to highlight the PointPillars/MonoDIS if it achieves at least 20\% improvements at corresponding item to its counterpart. Data is collected from nuScenes~\cite{nuscenes}. }
\centering
\resizebox{\linewidth}{!}{
\begin{tabular}{lccccccccccc}
\toprule
~ & Barrier & Bicycle & Bus & Car & Constr. Veh. & Motorcycle & Pedestrian & Traffic Cone & Trailer & Truck & Mean\\ 
\midrule
AP & 
38.9/\textcolor{blue}{51.1} & 
1.1/\textcolor{blue}{24.5} & 
\textcolor{red}{28.2}/18.8 & 
\textcolor{red}{68.4}/47.8 & 
4.1/\textcolor{blue}{7.4} & 
27.4/29.0 & 
\textcolor{red}{59.7}/37.0 & 
30.8/\textcolor{blue}{48.7} &  
\textcolor{red}{23.4}/17.6 & 
23.0/22.0  &  
30.5/30.4  \\

ATE & 
0.71/\textcolor{blue}{0.53} & 
\textcolor{red}{0.31}/0.71 & 
\textcolor{red}{0.56}/0.84 & 
\textcolor{red}{0.28}/0.61 & 
0.89/1.03 & 
\textcolor{red}{0.36}/0.66 & 
\textcolor{red}{0.28}/0.70 &
\textcolor{red}{0.40}/0.50 &
0.89/1.03 & 
\textcolor{red}{0.49}/0.78 & 
\textcolor{red}{0.52}/0.74 \\

ASE & 
0.30/0.29 & 
0.32/0.30 & 
0.20/0.19 & 
0.16/0.15 & 
0.49/\textcolor{blue}{0.39} &
0.29/0.24 &
0.31/0.31 &
0.39/0.36 &
0.20/0.20 &
0.23/0.20 &
0.29/0.26 \\

AOE &
\textcolor{red}{0.08}/0.15 &
\textcolor{red}{0.54}/1.04 & 
0.25/\textcolor{blue}{0.12} & 
0.20/\textcolor{blue}{0.07} & 
1.26/\textcolor{blue}{0.89} & 
0.79/\textcolor{blue}{0.51} & 
\textcolor{red}{0.37}/1.27 & 
-/- & 
0.83/0.78 & 
0.18/\textcolor{blue}{0.08} & 
0.50/0.55  \\

AVE &
-/- & 
\textcolor{red}{0.43}/0.93 & 
\textcolor{red}{0.42}/2.86 &
\textcolor{red}{0.24}/1.78 & 
\textcolor{red}{0.11}/0.38 & 
\textcolor{red}{0.63}/3.15 & 
\textcolor{red}{0.25}/0.89 & 
-/- & 
\textcolor{red}{0.20}/0.64 & 
\textcolor{red}{0.25}/1.80 & 
\textcolor{red}{0.32}/1.55 \\

AAE &
-/- & 
0.68/\textcolor{blue}{0.01} &
0.34/0.30 &
0.36/\textcolor{blue}{0.12} &
0.15/0.15 &
0.64/\textcolor{blue}{0.02} &
0.16/0.18 &
-/- &
0.21/0.15 &
0.41/\textcolor{blue}{0.14} &
0.37/\textcolor{blue}{0.13} \\
\bottomrule
\end{tabular}}
\label{table:monodis_pointpillars}
\end{table*}

\subsubsection{Metrics and Applications}

As introduced in Section~\ref{sec:dataset}, the metrics used in 3D detection are mainly derived from AP. In particular, KITTI 3D and Waymo Open use 3D/BEV IoU as criterion to distinguish a prediction among TP or FP, and then compute the precision-recall curve. However, the 3D/BEV IoU is sensitive to 3D position and goes from 1 to 0 quickly, which makes it difficult to detect far-away objects, especially for image-based methods~\cite{monodle,oftnet,nuscenes}. However, these far-away objects are less important than the near ones in the autonomous driving scenarios, and the metric should not so sensitive to these samples. In contrast, nuScenes use center distance as criterion to distinguish among TP and FP, and  evaluate each item of the objects separately.  Under this setting, image-based methods achieve higher accuracy, even surpassing LiDAR-based methods in some cases \cite{nuscenes}. However, this metric still applies unified standards for all samples, instead of treating different samples differently \cite{cityscapes3d,metricsde}.

\begin{table}[!t]
\caption{Performances of PointPillars \cite{pointpillars} and MonoDIS \cite{monodis}. For KITTI, we show the ${\rm AP}|_{R40}$ of the Car category on the \emph{test} set. For nuScenes, we show the mAP and NDS on the \emph{test} set. }
\centering
\begin{tabular}{lccccc}
\toprule
\multirow{2}{*}{~} & \multicolumn{3}{c}{KITTI} & \multicolumn{2}{c}{nuScenes} \\
\cmidrule(lr){2-4} \cmidrule(lr){5-6}  
~ &  Easy & Moderate & Hard & mAP & NDS\\ 
\midrule
PointPillars~\cite{pointpillars} & 79.05 & 74.99 & 68.30 & 30.5 & 45.3 \\
MonoDIS~\cite{monodis} & 16.54 & 12.97 & 11.04 & 30.4 & 38.4 \\
\bottomrule
\end{tabular}
\label{table:monodis_pointpillars_overall}
\end{table}

Furthermore, we should note that different types of detection errors bring different potential hazards to practical applications. For example, it is more important to provide a prediction instead of missing it, even if the localization accuracy is not so accurate. However, for existing AP-based metrics, the penalty of giving a FP is greater than that of missing a TP (both cases have the same recall, but the former has a lower precision, see Equation \ref{equ:recall&precision} for the definition of recall and precision).

In summary, in addition to meet the basic requirements, we believe the ideal metric for the 3D detection task in autonomous driving applications should have the following two features:
(i) treat the objects at different distances differently and focus more on the near objects ({\it e.g.} reduce the weights or the criterion for the far-away objects in evaluation);
(ii) treat different types of errors differently ({\it e.g.} higher penalty for the missing than mislocalization).

\subsubsection{Accuracy and Speed}
The inference speed in the 3D detection task is equally important to the accuracy for autonomous driving applications. However, so far, most of the research works only focused on the accuracy of predictions. For instance, pseudo-LiDAR-based methods achieve some gains in performance, but they also introduce extra computational overhead because they use an auxiliary network to estimate the depth maps. In particular, the most commonly used depth estimators~\cite{dorn,pspnet} in pseudo-LiDAR models take around 400 ms to compute the depth map for a standard KITTI 3D image. This latency will cause about 4 meters shift for a vehicle with a speed of 36 km/h. Although the actual situation can not be modeled so simply, this is just an example to illustrate the importance of algorithms' speed.

Additionally, the research community around 3D object detection is not so well established yet as compared to those studying some fundamental CV tasks such as 2D detection or semantic segmentation. Consequently, standard evaluation protocols are less mature and it is difficult to force all methods to use the same backbones or frameworks for a fair comparison. In this respect, evaluating both accuracy and speed represent a necessary step forward for comparing different methods, similar to what the 2D detection community did a few years ago.

\subsubsection{Image-based Methods and LiDAR-based Methods}
\label{sec:image_vs_lidar}
Another main branch of the 3D object detection task is the LiDAR-based methods. Here we discuss the strengths and weaknesses of the image-based methods and LiDAR-based methods.

\noindent
{\bf Overall Performances on KITTI 3D and nuScenes}
In this section, we choose two representative methods\footnote[2]{We choose MonoDIS and PointPillar as examples because they are con-current and also adopted in the nuScenes's official report \cite{nuscenes}. More recent methods with better performance can be found in our website.} to deeply analyze the features of LiDAR-based methods and image-based methods. As shown in Table~\ref{table:monodis_pointpillars_overall}, there is a huge gap between the performances of PointPillars \cite{pointpillars} and MonoDIS \cite{monodis} on KITTI 3D, while they have a similar accuracy on nuScenes. This is a common phenomenon for existing algorithms, which is mainly caused by the following three reasons:
(i) The resolutions of LiDAR signals are different. KITTI 3D uses a 64-beam LiDAR to capture the objects, while a 32-beam LiDAR is adopted in nuScenes when data collecting.
(ii) The objects of interest are different.  KITTI 3D mainly focuses on the Car category, while nuScenes averages the performance of ten classes, including some small or holed objects that are not easily captured by LiDAR signals.
(iii) The metrics are different. The metric used in nuScenes disengages the objects and reduces the criterion of localization accuracy, which is the primary error type for image-based methods. In particular, nuScenes averages the AP under different distance thresholds, {\it i.e.} $\mathbb{D}=\{0.5,1,2,4\}$ meters, and the average distance error (ATE in Table \ref{table:monodis_pointpillars}) of MonoDIS is about 0.7m which is okay for 1.0m, 2.0m, and 4.0m thresholds.

\noindent
{\bf Detailed Performances Analysis on nuScenes}
nuScenes decouples detection and reports the accuracy for each item individually. This design allows us to analyse the strengths and drawbacks of image-based methods in detail. The following observations can be made according to the results in Table~\ref{table:monodis_pointpillars}:
(i) Although the mAP of the two algorithms is very close (30.5 {\it v.s.} 30.4), they show different patterns in class-wise evaluation. MonoDIS is good at holed objects ({\it e.g.} Barrier and Bicycle) and thin objects ({\it e.g.} Traffic Cone), while PointPillars shows a higher accuracy on large objects ({\it e.g.} Bus and Car).
(ii) In most cases, PointPillars has a lower ATE, which means it can estimate the location of objects more accurately.
(iii) The size of objects estimated from MonoDIS is slightly better than that from PointPillars.
(iv) Benefiting from the accurate spatial information provided by LiDAR signals, PointPillars can accurately estimate the instantaneous velocity of the objects.
(v) MonoDIS shows a better ability to recognize the attributes of objects ({\it e.g.} whether a car is stopped or moving), which is an important feature for autonomous driving.

\begin{table}[!t]
\caption{The relative mAP changes of MonoDIS and PointPillars on the subsets of the nuScenes \emph{validation} set.}
\centering
\begin{tabular}{lcccccc}
\toprule
~  & Singapore & Rain & Night\\ 
\midrule
PointPillars~\cite{pointpillars} & - 1\% & - 6\% & -36\%  \\
MonoDIS~\cite{monodis} & - 8\% & + 3\% & -58\%  \\
\bottomrule
\end{tabular}
\label{table:robustness}
\end{table}

\noindent
{\bf Generalization}
nuScenes provides the performance changes of PointPillars and MonoDIS on subsets of validation set (shown in Table~\ref{table:robustness}), which can be used to analyze the robustness of these methods. We can observe a small performance drop on the Singapore split for MonoDIS. This indicates that the change of data distribution will affect the accuracy of monocular-based methods (mainly caused by the biased depth estimation). Besides, although the performance of PointPillars in the rainy split has only a slight decrease, it may be worse in practice because some frames in this split are not ongoing rainfall~\cite{nuscenes}. The biggest challenge is the night data, and we can see both MonoDIS and PointPillars experience a significant performance drop. Furthermore, MonoDIS has a more significant performance decrease than PointPillars, which may indicate that image-based methods are more sensitive to poor lighting.

\subsubsection{Image, pseudo-LiDAR, and LiDAR}
\label{supp:dis_pseudolidar}
\revision{As the bridge to link the RGB images and LiDAR points, pseudo-LiDAR representation plays an important role in image-based 3D detection and draws lots of attention, especially in 2019 - 2021. Specifically, compared with real LiDAR points, applying pseudo-LiDAR in 3D detection is a much cheaper approach, avoiding the involvement of expensive and cumbersome LiDAR equipment. Besides, due to the high resolution of RGB images, the generated pseudo-LiDAR typically has a higher density than real LiDAR, making it easier for models based on such data to capture small objects. However, due to the depth information is estimated, instead of measured, the overall detection accuracy significantly lags behind the real LiDAR-based models.}

\revision{Besides, from the perspective of the image-based model, pseudo-LiDAR representation provides a friendly interface to leverage the advanced techniques of LiDAR-based 3D object detection. Specifically, image-based 3D detection models are not as well-developed as the LiDAR-based models, and the detection accuracy is only 2.9 AP (GS3D \cite{gs3d}, the best monocular model before pseudo-LiDAR) in KITTI 3D dataset. The pseudo-LiDAR representation allows us to take the techniques in the LiDAR-based 3D detection community, facilitating the detection accuracy of image models to achieve about 13 AP within one year. Meanwhile, also noted that, compared with other image-based models, pseudo-LiDAR-based models also have some disadvantages, mainly caused by the additional phase to generate pseudo-LiDAR signals. This step not only introduces additional cost (for both training and inference) but also divides the detection model into two independent parts (depth estimation and 3D detection) and the stage-by-stage training may lead to sub-optimal results. Although some works \cite{pseudolidare2e,isplneeded} attempt to design end-to-end pseudo-LiDAR-based models, this research line has not been well-developed, and more efficient and effective detection pipelines, such as the BEV pipeline \cite{bevdet} or the DETR pipeline \cite{detr3d}, are more popular for now.}